\documentclass[acmsmall, screen, review=false]{acmart}
\AtBeginDocument{%
  \providecommand\BibTeX{{%
    \normalfont B\kern-0.5em{\scshape i\kern-0.25em b}\kern-0.8em\TeX}}}

\usepackage{xspace}
\usepackage{footnote}
\usepackage{tabularx} % table width
\usepackage{siunitx}
\usepackage{microtype}
\usepackage{amsmath}
\usepackage{comment}
\usepackage{xcolor}
\usepackage{enumitem}
\DeclareMathOperator*{\argmax}{argmax}

\newcommand{\name}{{\textsc{Virtuoso}}\xspace}
\newcommand{\namemeaning}{Just like a \name is a person who has exceptional skill, expertise, or talent at some endeavor, we believe our system demonstrates such skill in configuring streaming video object detection on embedded or mobile devices.\xspace}
\newcommand{\eg}{{e.g.,}\xspace}
\newcommand{\ie}{{i.e.,}\xspace}
\newcommand{\et}{\textit{et al.}\xspace}

\newif\ifdraft

\newcommand{\ran}[1]{\textcolor{orange}{\textbf{RAN:} #1}}

\begin{document}

% \title{Benchmarking Video Object Detection Systems on Embedded Devices under Resource Contention}
% \title{\name: Efficient and Adaptive Video Object Detection Systems for Mobiles}
\title{\name: Video-based Intelligence for real-time tuning on SOCs}

\author{Jayoung Lee$^\psi$, Pengcheng Wang$^\psi$, Ran Xu$^\psi$, Venkat Dasari$^\lambda$, Noah Weston$^\lambda$, Yin Li$^\delta$, Saurabh Bagchi$^\psi$, Somali Chaterji$^\psi$
% $\alpha$: Equal contribution \\ 
\newline
\newline
$\psi$: Purdue University, $\lambda$: Army Research Lab, $\delta$: U of Wisconsin at Madison}
\date{}

\begin{abstract}
% Ran Xu (10/17/2021):
% Enhance option 1: take the memory consumption into consideration
% Enhance option 2: take the model switching into consideration
% Enhance option 3: Add ApproxDet into our benchmark
% Enhance option 4: Adapt to resource contention as well
Efficient and adaptive computer vision systems have been proposed to make computer vision tasks, such as image classification and object detection, optimized for embedded or mobile devices. These solutions, quite recent in their origin, focus on optimizing the model (a deep neural network, DNN) or the system by designing an adaptive system with approximation knobs. In spite of several recent efforts, we show that existing solutions suffer from two major drawbacks. \textit{First}, the system does not consider energy consumption of the models while making a decision on which model to run. \textit{Second}, the evaluation does not consider the practical scenario of contention on the device, due to other co-resident workloads.
In this work, we propose an efficient and adaptive video object detection system --- \name, which is jointly optimized for accuracy, energy efficiency, and latency. Underlying \name is a multi-branch execution kernel that is capable of running at different operating points in the accuracy-energy-latency axes, and a lightweight runtime scheduler to select the best fit execution branch to satisfy the user requirement. We position this work as a first step in understanding the suitability of various object detection kernels on embedded boards in the accuracy-latency-energy axes, opening the door for much farther development in solutions customized to embedded systems and for benchmarking such solutions. 
%YL112321: modified to make things more consistent. "object detection heads" -> "object detection kernels" 
%
% Calculation of "286 fps" and "50 times": 
% --> "286" fps for fastest possible branch using d0. 45 times faster latency for D3.
% Details of 97.8% energy reduction: From D3 model branch.
% For EfficientDet D3, the original model performs at 63.9\% accuracy with 245.3 msec latency and 10.8 J energy consumption per frame for power mode 0. This is reduced down to 5.4 msec and 0.3 J per frame which is 45 times faster, and 97.2\% more energy efficient at the cost of lower accuracy of 39.5\% 
With this, \name is able to achieve up to 286 FPS on the NVIDIA Jetson AGX Xavier board, which is up to 45 times faster than the baseline EfficientDet D3 and 15 times faster than the baseline EfficientDet D0.
% \somali{and XXX times faster than the baseline EfficientDet D0}. 
%JL110521: Addressed
%SC101921: which is what?
In addition, we also observe up to 97.2\% energy reduction using \name compared to the baseline YOLO (v3) --- a widely used object detector designed for mobiles.
%SC102121: Which baseline? Be specific. 
% Jay : 10/17/2021 For yolo, yolov3 is only available in arxiv, and all the citations are for the arxiv article as well. I have included the conference for the first version of yolo.
%YL102221: Is there a reason why we are picking the vanilla yolo (instead of a later version) as the baseline?
To fairly compare with \name, we benchmark 15 
%SC102121: don't say 10+, give the exact number. 
state-of-the-art or widely used protocols, including Faster R-CNN (FRCNN) [NeurIPS'15], YOLO v3 [CVPR'16], SSD [ECCV'16], EfficientDet [CVPR'20], SELSA [ICCV'19], MEGA [CVPR'20], REPP [IROS'20], FastAdapt [EMDL'21], and our in-house adaptive variants of FRCNN+, YOLO+, SSD+, and EfficientDet+ (our variants have enhanced efficiency for mobiles).
%YL102221: It will be hard to argue everything here is SOTA. I would just use "latest protocols"
%
With this comprehensive benchmark, \name has shown superiority to all the above protocols, leading the accuracy frontier at every efficiency level on NVIDIA Jetson mobile GPUs. 
%YL102221: Can we say a bit more about what type of devices we evaluted on?
% Effdet D3 compared against FRCNN/YOLO
Specifically, \name has achieved an accuracy of 63.9\%, which is more than 10\% higher than some of the popular object detection models, FRCNN at 51.1\%, and YOLO at 49.5\%.
\end{abstract}

\maketitle
%JL110521: Slightly cleaned up some of the old comments. Left others just in case.
\section{Introduction}\label{sec:intro}

Video analytic systems have seen widespread success in various domains, ranging from computationally heavy tasks such as recognizing faces for surveillance, to mobile applications such as detecting objects for mobile-based augmented reality (AR)~\cite{rohan2019convolutional, apicharttrisorn2019frugal, rao2017mobile}, and to real-time systems such as localizing pedestrians and cars for autonomous driving~\cite{feng2020deep, arnold2019survey, li2019gs3d}. A key function shared by these applications is the ability to detect objects in videos.
There is a growing number of use cases for performing such video object detection on {\em mobile devices}. These devices are increasingly equipped with mobile GPUs, albeit they are much weaker computationally than those on server-class machines. 

Nevertheless, there is an impetus to perform the video processing in (near) real-time on the streaming (video) content, on the device itself\footnote{For convenience we will often use the shorthand ``device" to refer to a ``mobile device".}. This is needed, say, to improve the user's immersive experience (\eg in AR/VR games) or to give high-confidence outputs from streaming videos (\eg for pedestrian recognition in autonomous driving). A typical operating point is processing each frame in 33 ms, which corresponds to the video stream rate of 30 frames per second (FPS). We find empirically that state-of-the-art (SOTA) protocols (designed for servers) when executed on mobile platforms significantly overshoot this latency margin. For example, MEGA~\cite{chen2020memory}, considered a SOTA solution, takes 253.4 msec per frame on NVIDIA AGX Xavier, when occupying all computing resources. 
% \pengcheng{TODOs for Jay: Fill in the XXX}
%JL110521: Addressed.
In response, a slew of efficient models and systems have been proposed to improve their efficiency or performance on mobile devices~\cite{tan2020efficientdet, sandler2018mobilenetv2, jiang2020learning, zhu2018towards, chen2018optimizing, liu2016ssd, redmon2018yolov3, zhang2018single}. %starting 5 years back and with continued robust activity today.
Another approach to make the streaming analytics feasible on mobile devices is to utilize both object detection and tracking, which is captured by the technique ``\textit{tracking-by-detection}''~\cite{bae2014robust, muller2018trackingnet}. 
% ``\textit{Tracking-by-detection}'' addresses the known limitation of object detection and tracking wherein factors such as occlusion, fast motion, or deformation of the object within the content could degrade the performance~\cite{wu2013online}. 
%
By interspersing multiple frames of tracking (light-weight compared to detection) with each frame where object detection is executed, the overall performance can be sped up. 
%SC110121: Basic background --- not needed in introduction, move to background. 
% This is because the tracker utilizes the detection results from the detector, not needing to learn all the variations of the object. Hence, most tracking algorithms are much faster than object detection.
%JL110521: Considered to be duplicate content. So keeping it commented here.
%
%
% Tracking-by-detection exposes several configuration knobs. 
%JL110521: Since we use efficiency knobs as one of our contributions, avoid using knobs.
The tracking-by-detection technique exposes several adaptation strategies.
%JL110521: Modified below sentence.
This can include selecting among a set of detectors and trackers and how to handle the trigger frequency between the detector and tracker.
%JL110521: Moved the below sentence here.
Empirically, we find that the execution time of an object tracker is 10X lower than an object detector.
A judicious choice of such adaptation strategies is needed to satisfy the real-time requirements for energy, latency, and accuracy, especially on mobile devices. %considering that other co-existing applications on them and their limited energy storage. 

%Despite the fact that ``\textit{tracking-by-detection}'' has some advantages, 
The stringent requirements on mobile devices at runtime expose another challenge for streaming video analytics: \textit{how to adapt (at runtime) to the dynamic user requirements and available resources on the device?} 
To address this problem, several recent works considered multi-branch solutions~\cite{fang2018nestdnn, xu2020approxdet, xu2019approxnet, lee2021benchmarking}. 
% \somali{there is no technical difference between multiple-branch and multiple-path so just retain one.}
% PC (11/02/2021): Only kept one
%JL110521: changed it to multi-branch to keep terminology consistent across the paper.
Such solutions include multiple execution kernels in a system and choose the optimal one {\em during runtime} to satisfy the user requirement of latency. %accuracy or latency. 
%YL112321: removed accuracy. I don't think there is such model to satisfy user specified accuracy.
%JL112421: Thank you, yes accuracy seems inappropriate here.
Yet, omitting features that control energy consumption is a major drawback of all these approaches. 
% Further, the execution backbone itself is rather outdated.
Further, compared to recent studies on efficient convolutional neural network (CNN) architectures, there has been limited studies and applications on efficient object detection solutions on embedded devices, along with the issue of using outdated feature extractors.
%SC110121: Above sentence is not defensible. People will say then just use the latest object detector and expose multiple branches from it. 
%JL110521: Reworded, tried to compare against CNN works itself without the object detection head. If to add citations, FBNet, EfficientNet, CSPNet, GhostNet could be some of the picks.

%YL102221: We should emphasis the key drawback (omitting energy consumption) before the secondary ones (outdated execution kernel). I also don't quite understand the argument that "they only consider latency requirements". ApproxDet, for example, considered both latency and accuracy. 
% PC (10/24/2021): Changed the order, and added accuracy requirement.
% PC (10/24/2021): One thing I want to discuss is that according to my understanding , prior work cannot satisfy accuracy requirement. For example, ApproxDet satisfy the latency requirement, at the same time it max the accuracy performance. But it seems cannot satisfy specific accuracy requirement, e.g. 60%, 50%. However, we also don't have the option to satisfy a specific accuracy requirement, neither. Maybe that's something we can improve.

% Ran (10/17/2021): Enhance option: add memory footprint back
In real applications, the change of video content, available resources on mobile devices, and manual control of the requirements make the dynamic adaptations even harder. The wide range of energy, latency, and accuracy requirement means the system needs to have the adaptability to a variety of scenarios.
%JL110521 : Changed above requirement order - eng -> lat -> acc
Thus, the design must satisfy the following prerequisites: (1) designing an execution kernel that is both energy and latency efficient for the mobile device, (2) analysis of the energy consumption and latency performance of all execution branches on mobile devices, and (3) an adaptive scheduler to make decisions at runtime to satisfy multiple requirements simultaneously. However, to our knowledge, no prior work has included all of these design innovations. 
% \somali{Check above sentence.}
%JL110521 : Slight modification.

% PC (11/01/2021): I think it is hard to say this for 1)
% considered the energy consumption of an object detection model on mobile devices, while optimizing the trade-off for latency or accuracy performance. 
%SC102121: So are we saying that the two other requirements are met by prior work?
%PC (10/24/2021): If we want to say that prior cannot satisfy latency requirement, one thing we can say is that they cannot meet some stringent requirements, for example 10 msec per frame.

Another dimension that has been ignored to date is how do these algorithms behave when there is resource contention on the device. Such resource contention can happen due to co-located applications running on the same device, an occurrence that is quite common because these are multi-purpose devices. Hence, say, when there is video detection going on, another computationally heavy task such as speech recognition may also be executing on the device.  
No prior work has evaluated latest video object detection systems, under varying resource contention on leading GPU-enabled mobile devices.
%

%YL102221: The transition between this paragraph and the previous work needs some work. It is not clear what is the challenge of adapting to varying requirements? "no prior work ..." was already covered by the previous paragraph. The mentioning of contention is also quite abrupt. 
% PC (10/25/2021): Added some challenges. But the transition do need more work to reorganize the paper. I have some plan and we can discuss on the Tuesday meeting.
%SC110121: Added

% PC (10/18/2021): Add the overall design figure
\begin{figure}[t]
    \centering
    \includegraphics[width=1\textwidth]{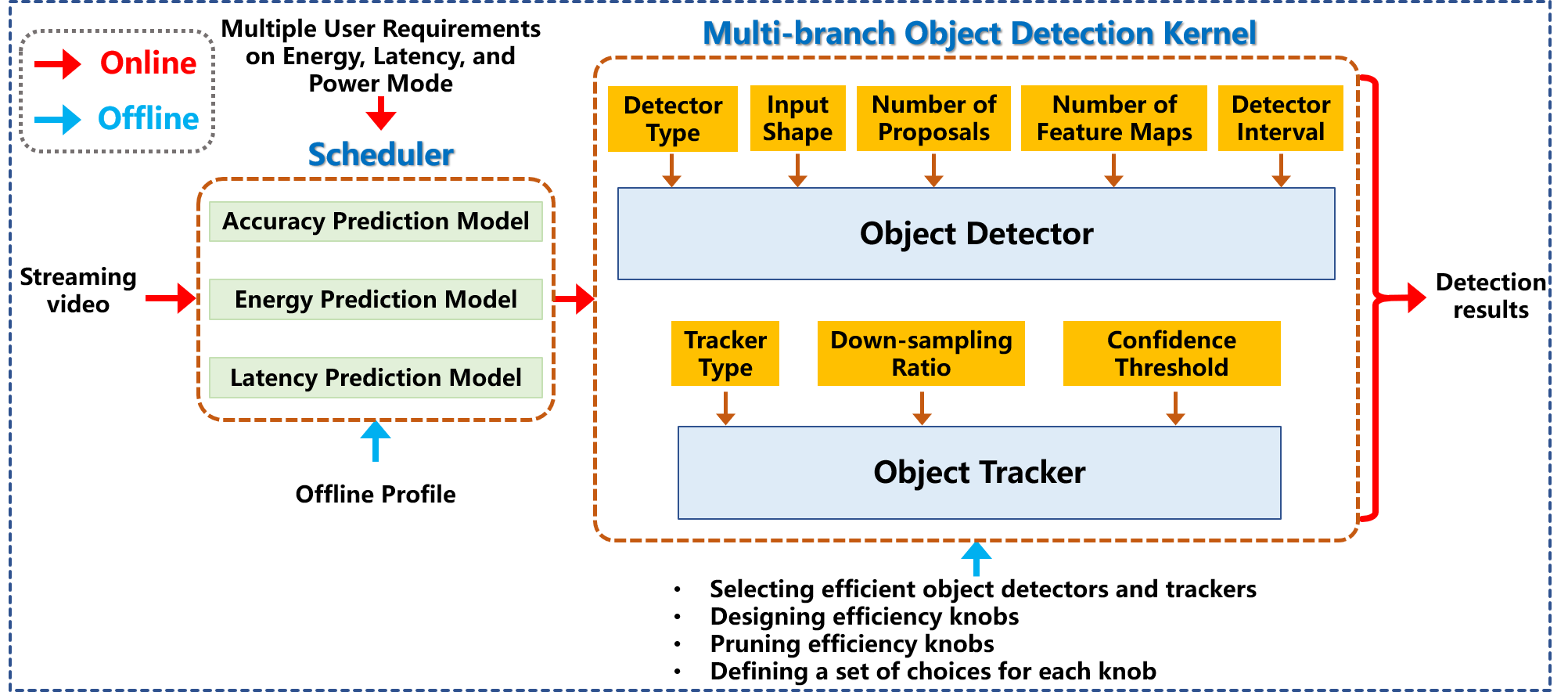}
    \caption{The illustration of \name, which includes two important components, a multi-branch object detection kernel and a scheduler. The multi-branch object detection kernel was achieved by exposing multiple tuning knobs for both object detector and object tracker. The scheduler can do latency, energy, and accuracy prediction at runtime to satisfy user requirements.}
    \label{fig:overall}
    % \ran{Polish: (1) ``Offline Profling''-->``Offline Profile'', (2) Scheduler block: ``Accuracy'', ``Energy'', and finally ``Latency'', (3)``Multiple User Requiremnts'' --> ``Multiple User Requirements on energy, latency and power mode'', (4) Add 2 more efficiency knobs -- Detector variant and number of feature maps, (5) ``Confidence Score Threshold'' --> ``Confidence Threshold'' (6) ``Selecting efficient object detectors and trackers'', (7) ``Designing efficiency knobs'', (8) ``Pruning efficiency knobs'', (9) ``for each knob (not knobs)''}
    % \pengcheng{addressed}
\end{figure}

% PC (10/18/2021): Add the summary of our proposed techniques. A overall design figure is needed for better visualization and illustration.
In this work, we present \name\footnote{\namemeaning}, which is customized for mobile devices under varying resource constraints, with additional efficiency knobs\footnote{We use the term ``efficiency knobs'' rather than the more common term ``tuning knobs'' as we are focusing on the performance metric of accuracy, normalized by latency or energy cost to achieve that accuracy, in other words, efficient accuracy.} for energy-aware adaptation.
%SC102121: we are adding memory consideration then? That will be helpful.
%PC (10/23/2021): Delete memory as it will not be included in this paper
\name selects the most efficient baseline object detectors EfficientDet~\cite{tan2020efficientdet} and SSD~\cite{liu2016ssd}, enhances them by integrating object trackers, and provides 155 execution branches by exposing 8 efficiency knobs within one system. Moreover, \name has a scheduler that can select the optimal branch during runtime to satisfy the energy and latency requirements at the same time, while maximizing the accuracy.
We also evaluated a total of 15 baseline models on 3 different embedded boards under 2 resource contention levels to better understand the energy, latency, and accuracy performance of latest models for video object detection. 

% PC (10/18/2021): Added brief description of our evaluation and results. But the current description is fragmented and we need a good story to put them together. XXXs are left for Jay to fill in.
% Jay 10/19: Addressed
We evaluate \name and the baselines using the ILSVRC 2015 VID datasets on 3 NVIDIA mobile GPUs of increasing compute capacity --- Jetson TX2, Xavier NX, and AGX Xavier, under 2 different power modes of AGX Xavier. %These embedded devices are stand-ins for high-end mobile devices with GPU. %
On each board and under each power mode, \name is the best solution that can automatically choose the optimal branch to satisfy the energy and latency requirements, and maximize the accuracy performance. 
% Jay 10/19 : For internal purpose.
% For lowest latency
% [[['EFFDETD0', -1, -1, 100, 'medianflow', 2, 30, 'agx0'], [0.4461, 0.3246, 3.4966745628999996, 0.2, 0]]]
% For lowest energy
% [[['EFFDETD0', -1, -1, 100, 'medianflow', 4, 30, 'agx2'], [0.4446, 0.3286, 4.5733791865, 0.122, 0]]]
With the multi-branch execution kernel, \name can adapt its energy efficiency within a wide range (span of 89X) and also latency in a wide range (span of 128X). Such wide ranges of energy efficiency and latency can satisfy various scenarios at runtime to optimize the functionality of our video object detection system. 
% The lowest latency that \name can achieve is 3.5 msec with a 32.5\% accuracy, the lowest energy consumption is 0.12 J per frame with accuracy of 32.9\%.
Given 1 J per frame energy requirement, \name can achieve 52.0\% accuracy, and given 33.3 msec latency budget, \name can achieve 60.1\% accuracy on an AGX Xavier board. 
% \pengcheng{Todos for Jay: fill in the XXX.}
%JL110521: Addressed.
%
% Jay 11/2: Changed to 33.3 msec latency budget, which seems more practical.
Additionally, when energy and latency do not have stringent limits, \name can achieve 63.9\% accuracy, which is more than 10\% higher than baseline models.

In this paper, our contributions are as follows.
\vspace{-3pt}
\begin{enumerate}[leftmargin=1em]
    % Ran (10/17/2021): enhance option -- adding memory constraint into our consideration
    \item We present \name, the best energy efficient and adaptive video object detection system for mobile devices. \name can dynamically adapt its runtime configurations based on the given user requirements. Our runtime scheduler solves the optimization problem for accuracy at any energy and latency level and thus can guarantee the Pareto optimal performance on resource-constrained devices. We are the only adaptive model doing this 3D (energy-latency-accuracy) tuning while handling resource contention.

    \item \name designs an efficient multi-branch execution kernel with a total of 8 different efficiency knobs, 
        which are optimized for both energy efficiency and latency on mobiles. This gives \name more flexibility to a wider range of available resources, span of 89X energy adapting space when baseline models cannot, and 4.5X times larger latency adapting space than the baseline FastAdapt~\cite{lee2021benchmarking}. \name can achieve 97\% lower energy consumption and up to 50 times faster execution time, compared to using the object detection kernel only.
    %YL112321: "the object detection kernel". Do you refer to the best object detection kernel here?
    %SC102121: It is vague what we mean by "original object detection model".
    % PC (10/25/2021): Changed it to "object detection kernel only"
    % Ran (10/17/2021): Usage of a tracker is one example approximation knob. Usage of SSD and EfficientDet is another example approximation knob. Thus they should not be emphasized in the contribution.
    %\name consists of an object detector and an object tracker with tuning knobs as the key component along with a dynamic scheduler that can change configurations of the framework real-time. \name utilizes SSD coupled with MobileNetv2, and EfficientDet as its detector backbone to be optimized for mobile performance. 
    \item We evaluate our proposed systems and 15 baselines (including 3 latest video object detection solutions and our enhanced FRCNN~\cite{ren2015faster}, YOLO~\cite{redmon2018yolov3}, SSD~\cite{liu2016ssd}, and EfficientDet~\cite{tan2020efficientdet} for both energy and latency on the embedded devices) on 3 NVIDIA Jetson embedded devices and under different GPU resource availability, energy efficiency constraint, and latency constraints.
    % Jay 11/2 : Added more content for power modes
    In addition, we also investigate and evaluate the impact of different device power modes, which has not been evaluated before. 
    %
    % \somali{has not been evaluated before is correct, right?}
    % PC (11/04/2021): I believe this is true.
    %Different power modes control the performance of the embedded device which is a feature that could further optimize the energy and latency performance.
    %SC: I commented the sentence above, redundant and added the one below.
    %
    Different power modes allow several configurations with different different CPU frequencies and number of CPU cores online.
    %SC: https://docs.nvidia.com/jetson/archives/l4t-archived/l4t-3243/index.html#page/Tegra%2520Linux%2520Driver%2520Package%2520Development%2520Guide%2Fpower_management_tx2_32.html%23wwpID0E01N0HA
    %
    We show accuracy superiority to baselines given any efficiency constraint.
    % \somali{Emphasize the power modes part a bit.}
    
\end{enumerate}

% SB (11/05/21): Add a paragraph saying: "The rest of the paper is organized as follows. XXX"
% Jay 11/9: Addressed in a very high-level content, from section 3~5.
The rest of the paper is organized as follows. In Section.~\ref{sec:tech}, we present our overall design for \name and its key components, such as the dynamic scheduler and efficiency knobs for handling various user requirements for latency or energy consumption. In Section.~\ref{sec:implementations}, we present the multi-branch kernel design and implementation of \name, along with different runtime environments for evaluation. Section.~\ref{sec:evaluation} consists of two parts: First, we evaluate the overall performance of \name. Next, we evaluate \name and its multi-branch kernels against other baselines.

\section{Related Work}\label{sec:related-work}

\noindent\textbf{Video object detection} seeks to locate object instances in video frames using bounding boxes and simultaneously classify the instance into target categories with their class probabilities. The most widely used detection models adopt CNNs, broken down into two parts: a backbone network that extracts features from images (\eg ResNet), and a detection network or head, which classifies object regions and refines the localization of the objects based on the extracted features (\eg Faster R-CNN or YOLO). The detection network can be further categorized into two-stage detectors~\cite{dai2016r, ren2015faster, tan2020efficientdet}, or single-stage detectors~\cite{liu2016ssd, redmon2018yolov3, zhang2018single}. One representative work of two-stage detectors is Faster R-CNN (FRCNN)~\cite{ren2015faster}, where plausible regions are proposed in the first stage, followed by decision refinement in the second. Specifically, CNNs extract image feature maps and feed them into Region Proposal Networks (RPN) to generate regions-of-interest (RoIs) in the first stage. Then, in the second stage, the RoI pooling layer combines the feature maps from convolutional layers and the proposals from the RPN together to generate proposal feature maps and provide these to the classifier network. On the other hand, YOLO and SSD are the representative works for single-stage detectors. These single-stage end-to-end detection solutions do not include the step of region proposal generation, rather, directly classify a dense set of pre-defined regions from the feature maps. One-stage detection models are usually easier to train and are more computationally efficient, but often suffer from lower accuracy, especially for mAP with high IoU thresholds.

A general trend in the object detection is to design deeper and more complex object detection networks in order to achieve higher accuracy such as in recent video object detection algorithms~\cite{wu2019sequence, chen2020memory, sabater2020robust, yao2020video, chen2020memory, zhu2017flow, deng2019relation}. 
There is ongoing research on pushing the accuracy further for video object detection tasks, for example frame aggregation~\cite{yao2020video, chen2020memory}, a technique that utilizes features from other frames during inference to enhance the detection results. SELSA~\cite{wu2019sequence} widens the window for selecting the frames for aggregation by not only selecting neighboring frames, but considering their semantic neighborhood. MEGA~\cite{chen2020memory} takes the work from SELSA one step further and adds global frame aggregation where frames from other videos, sharing semantic similarity, are also taken into account. While these techniques are performed during the runtime of the inference task, REPP~\cite{sabater2020robust} reuses the detection output from a baseline model to further post-process the detection output to enhance the detection results after the analysis of a video. Other works make use of optical flow~\cite{zhu2017flow}, or techniques such as knowledge distillation~\cite{deng2019relation}. %that have not been commonly deployed in video object detection. 
However, these advancements in accuracy do not necessarily target making these algorithms more efficient in terms of the network size, energy consumption, and the latency of the detection task.
% Ran (10/26/2021): Can we use an energy budget instead of the latency budget?
% PC (10/26/2021): Added energy budget, but it is kind of hard to give a specific budget number. So, I used ``limited battery''
% Ran (10/27/2021): Great!

% Ran (10/26/2021): Any work on improving the energy efficiency?
% GhostNet, EfficientDet, MobileNet,
% PC (10/26/2021): They haven't mentioned energy efficiency explicitly, but they improve the computational efficiency which is related to energy efficiency. I will say that at the end of the paragraph.
% Ran (10/27/2021): Great!
Several studies have been conducted to optimize accuracy and latency for video object detection tasks. Feichtenhofer~\et~\cite{feichtenhofer2017detect} combine an object detector and an object tracker to create a joint design that is trained and deployed in an end-to-end fashion so that a light-weight tracker could speed up the process of a detector-only design. Jiang \et~\cite{jiang2020learning} use an LSTM module to propagate the high-level features across frames to reduce the computation cost resulting from the optical flow technique that captures the temporal information in the video. A ``key frame'' concept is used in~\cite{zhu2018towards} to effectively group adjacent frames with similar features, thus saving redundant computation costs. Chen \et~\cite{chen2018optimizing} also utilize the concept of ``key frame'', to adaptively schedule the computation path to sparsely spread out operations with high computation cost. However, most studies that tackle the optimization challenge between accuracy and latency still focus on server-class GPUs, which are much more powerful than mobile or embedded devices~\footnote{We use the terms ``mobile device'' and ``embedded device'' synonymously.}. In many real-world object detection tasks, the task has to be carried out in a real-time fashion on a computationally constrained platform, such as a mobile device. In such cases, video object detection becomes challenging because of the resource constraints and the stringent energy budget (limited battery) and latency budget (30--50 msec/frame) for acceptable video quality. 

Some recent approaches take the real-world computation constraint into account and design efficient backbones that are specifically designed to reduce the computation cost. MobileNetV2~\cite{sandler2018mobilenetv2} uses an inverted residual block to reduce the number of computations, and thus improves computational efficiency. AdaScale~\cite{chin2019adascale} makes use of the content information of videos to dynamically re-scale the images to lower resolution, and at the same time, achieve better accuracy. GhostNet~\cite{han2020ghostnet} uses a ``ghost module'' to reuse some of the features from the feature map to reduce the computational cost. These are examples that utilize a human-crafted component to optimize the model. On the other hand, model architectures can be automatically optimized using a neural architecture search (NAS) technique. NAS-based models~\cite{tan2019mnasnet, wu2019fbnet, tan2019efficientnet} pre-define the blocks or layers that will be used to construct the network, and search through combinations and connections of the pre-defined components. EfficientDet~\cite{tan2020efficientdet} introduces reinforcement learning (RL) based-NAS for a light-weight network. Instead of modifying or creating a new network design, some solutions add adaptive components to the pipeline, such as using a dynamic pipeline adapting to content at runtime~\cite{fang2018nestdnn, xu2020approxdet}. However, all of the aforementioned studies focus on the network design, which perform the object detection task in limited scenarios. Even though these works have improved the computational efficiency, they still require further development and improvement to be deployed for an real-time dynamic environment on mobile devices with changing energy and latency requirements. For example, they may require significant feature engineering to fit the specialized capabilities of the mobile GPUs as done by our prior work~\cite{ghoshal2015ensemble}.
%
%JL110521: Gave some thoughts on whether to change the' Benchmarking' below to Evaluating, but seemed benchmarking is slightly better since we are considering related works. Will change to evaluating if this 'presents' our paper to look like a benchmark paper.

\noindent\textbf{Benchmarking video object detection works on embedded devices:}
With the rise of video object detection, coupled with the popularity of edge computing in recent years, video object detection tasks have been pushed to the edge/embedded devices where the data is generated. %Performing these works at the edge has lots of benefit, including but not limited to: lower cost in data transmission, lower end-to-end latency, more security for private data, etc. 
%Edge computing is generally performed on embedded devices with limited computational power and thus users should have a better understanding on them by looking at a thorough benchmarking experiment. 
%Alyamkin \et~\cite{alyamkin2019low} have provided the vision on low power computer vision works.
%https://arxiv.org/pdf/1904.07714.pdf
%https://ieeexplore.ieee.org/abstract/document/6114206
MEVBench~\cite{clemons2011mevbench} has provided a benchmark suite for a range of mobile vision applications such as face detection, object tracking, and feature extraction. However, none of SOTA works or latest devices have been used, and its evaluation is not on the GPU, which is the \textit{de facto} hardware for DNN based computer vision works. AIoT bench~\cite{luo2018aiot} also provides an AI tasks' benchmark suite based on Android and Raspberry-Pi, and covers different frameworks like TensorFlow and Caffe2. Nevertheless, there are no SOTA models in it and no evaluation is presented. Qasaimeh \et~\cite{qasaimeh2021benchmarking} has conducted benchmarks of accuracy, latency, and energy on a wide range of vision kernels and neural networks on multiple embedded devices, \ie ARM57 CPU, Nvidia Jetson TX2, and Xilinx ZCU102 FPGA. However, it only includes one GPU-enabled device and thus is not comprehensive since embedded devices with GPUs are very common nowadays. Also it has not included the SOTA models and does not focus on video object detection, which is \name's focus. Also, lots of video object detection solutions are designed by including multiple tuning knobs, Qasaimeh \et~\cite{ qasaimeh2021benchmarking} has not shown the accuracy-efficiency tradeoff on embedded devices. Buckler \et~\cite{buckler2017reconfiguring} take a more detailed look at the kernels of the image signal processing (ISP) pipeline, \eg it does a detailed investigation of how many stages of an ISP pipeline should be used, what algorithm the image sensor should use, and the quantization of the ADC. Consequently, it is less complete in terms of its coverage of the detection kernels. It covers only one model for object detection, Faster R-CNN, which we also cover. They do not measure power consumption, but use analytical formulae and simulations. Euphrates~\cite{zhu2018euphrates} optimizes the interaction between the ISP and the CNN in the CV pipeline. This also does the optimization in a SoC architecture specific manner. However, it does not focus on the video object detection task.
%To conclude, none of the prior work has done thorough benchmark on the latency, accuracy, and energy performance of SOTA video object detection models/systems on cutting edge mobile devices with GPU modules. 
%
% Ran (10/26/2021): The conclusion here is a little strange. Do we understand better with our results? Can we modify these claims for our benchmarks?
% PC (10/26/2021): I rephrased some. May I know do you feel all the points here are strange or any one of them is strange?
% My thoughts is that we should need to write better observation and insights based on the benchmark results. For example: for point (2), AGX achieve 2X faster latency compared with TX2.  
% Ran (10/27/2021): AGX and TX2 are two stand-in devices for mobiles. No need to compare them and no value of showing the relative power between them. 
A comprehensive benchmark is important to understand the advantages and disadvantages of different efficient and adaptive models, \ie how accurate they are given an efficiency requirement and how much these models can adapt in terms of efficiency.
%
% Ran (10/27/2021): commented because we do not evaluate devices and the background should not mention the detailed tech, \eg tunning knobs.
%(2) the computational power and feature of popular GPU enabled mobile devices, (3) the effectiveness of different tuning knobs in video object detection systems. 
\section{Techniques}\label{sec:tech}

We now present the techniques used in the design and implementation of \name. To achieve high energy efficiency and low latency on embedded devices, we first propose a collection of efficiency knobs (Sec.~\ref{sec:efficiency_knobs}) and an efficient multi-branch object detection kernel (Sec.~\ref{sec:multi_branch}). Our system combines the efficiency knobs and is capable of running at different operating points in the accuracy-energy-latency axes through its multiple execution branches. We then propose our runtime scheduler to solve the constrained optimization problem at any energy efficiency or latency requirements (Sec.~\ref{sec:secheduler}). Finally, we design a tool to generate synthetic resource contention to benchmark each model under resource constraint scenarios (Sec.~\ref{sec:CG}).

\subsection{Efficiency knobs for Object Detection Models}\label{sec:efficiency_knobs}

% Evidence for "20 times": For SSD detector latency is 65 msec, while tracker latency is around 3.1 msec on average.
% Ran (10/28/2021): [Suggestion for paper version 3] 20 times is on latency efficiency. Can we use energy data, i.e., how much energy reduction of an object tracker compared to an object detector.
To effectively make object detection backbones both energy and time efficient, the \textit{tracking-by-detection} technique is one of the most common methods for video object detection. Particularly, considering a video as a sequence of consecutive frames, we define Group-of-Frames (GoF) as a collection of consecutive frames in which we apply the computationally expensive object detector to the first frame, and apply the light-weight object tracker, to the remaining frames. An object tracker is highly efficient since it is much cheaper in terms of
%SC011121: light-weight is an adjective and without the hyphen it is not an adj, e.g., light-weight tracker.
computation, with at least 20 times better latency performance than an object detector (from our results). However, it relies on the relationship between the current frame and the past frame, and the detection results of the past frame. Thus, an object tracker, in spite of being more efficient, cannot run without an object detector. The latter provides a calibrated detection result on every first frame of a GoF. %For context, we now introduce the following efficient methods which we address as ``efficiency knobs'', each coupled with either the object detector, or the object tracker. 
% \jay{Try to unify the terminology for 'knob' as efficiency knob'. There are currently efficient methods, knobs, approximation knobs, tuning knobs and so on.}
% Ran (10/28/2021): agreed.

%\jay{Decision after 10/27 : use object detector backbone / multi-branch object detection kernel}
% Ran (10/28/2021): update all the terms in this subsection.
% Jay 10/28 : added subsubsections to distinguish the knobs for detector and tracker
\subsubsection{Efficiency Knobs for the Object Detector} \hfill\\ % Force a new line

\noindent \textbf{Object Detection Backbone}:
We select a total of four different object detectors to perform the detection --- EfficientDet, SSD, FRCNN, and YOLO, and call them ``Object Detection Backbones''. %Each object detection backbone has its focus for the optimization, for example, EfficientDet and SSD are more focused on the efficiency on embedded devices. 
Switching among these kernels can be used as the primary adaptation strategy based on the users' requirements.
Particularly, EfficientDet is a family of object detectors that are scaled up with different scaling factors, starting from the base model D0. Among the 8 variants of EfficientDet (D0-D7), we experimentally find that model variants with larger scale than D3 fail to run on our embedded devices. Thus, we select EfficientDet D0 and D3 as the most light-weight, and heavy-weight ones, among all executable variants. These variants within the same object detector family enable the tradeoff with respect to accuracy, energy efficiency, and latency. For simplicity, an object detector backbone refers to an object detector, or a particular variant, \eg EfficientDet D0 or D3.

% Jay 10/28 : Comment/removed after decision.
% \noindent \textbf{Object Detection Kernel Variants}
% EfficientDet is a family of models that are scaled up with different scaling factors starting from the base model D0. Among the 8 variants of EfficientDet (D0-D7), we experimentally found that model variants with larger scale than D3 failed to run on our embedded devices. Thus we select EfficientDet D0 and D3 as kernel candidates for being the most light-weight, and heavy-weight model among the executable variants. These variants within the same kernel provides us with more flexibility on handling the trade-off between accuracy and efficiency.

\noindent \textbf{Input Image Resolution for the Object Detector Backbone}:
Given the different object detector backbones, we further consider additional efficiency knobs. %to further enhance the efficiency. 
First is the resolution of the input image fed into the detector. Each detector backbone comes with a pre-defined image resolution that can be processed through the neural network, and all input images are resized to the pre-defined resolution as the first step. We modify the input layer of the object detector backbone to accept images with different resolutions. This is possible as the backbones are full convolutional. Feeding a smaller-scaled image results in both less energy consumption and less computational overhead, translating to a more efficient model. %thus results in higher energy and time efficiency, at the possible drawback of loss of information or accuracy performance.

\noindent \textbf{Number of Proposals in the Object Detector Backbone}:
FRCNN is a two-stage object detector. The first stage is a Region Proposal Network (RPN) to process the feature map output from the feature extractor and return a pre-defined number of object candidates in the feature map. We implement number of proposals as the efficiency knob in the RPN to modify the number of output object candidates. The number of object candidates is directly related with the computation in the second stage of the detector. 
%SC110121: The \textit{number} of object candidates \textit{are} directly related with the computation in the second stage of the detector. [number can't go with are -- edited several slips like this so bringing this to your attention]
Thus, we are able to leverage the accuracy vs. efficiency tradeoff by modifying the number of proposals. This knob is only available for two-stage methods.

\noindent \textbf{Number of Feature Maps in the Object Detector Backbone}:
To further engineer our object detector backbone, we explore the MnasFPN~\cite{chen2020mnasfpn} feature pyramid in the SSD detector that is used to concatenate feature maps. 
%
% SB (11/7/21): There is an ordering to these feature maps. 1 is coarse grained and 2 is finer than that and so on, right? If so, that needs to be brought out.
% Jay 11/9 : Addressed below.
MnasFPN concatenates a total of four feature maps, responsible for detecting objects in different scales. The first feature map has the largest feature map size, and is responsible for detecting objects at a finer granularity. Following the path down the feature pyramid from the first feature map, the following feature maps are compressed gradually, and are used to detect larger objects.
We explore the tradeoff of the accuracy versus efficiency by using different combinations of the four feature maps, such as [1, 2, 3], [2, 4], and so on, where 1 to 4 stand for the four feature maps. 
%SC110121: tradeoff is the word; stick to it -- I see trade-off, trade off -- minor but pointing this out in this almost final pass so I don't have to edit again.

% Jay 10/28 : added subsubsections to distinguish the knobs for detector and tracker
\subsubsection{Efficiency Knobs for the Object Tracker} \hfill\\ % Force a new line

\noindent \textbf{Object Tracker}:
Another major component of our efficient design is the object tracker. Similar to different object detector backbones, we also utilize multiple object trackers. %to enable the option of varying focuses on either efficient performance or accuracy. 
A total of four object trackers, MedianFlow~\cite{kalal2010forward}, KCF~\cite{henriques2014high}, CSRT~\cite{lukezic2017discriminative}, and OpticalFlow~\cite{kale2015moving}, are utilized and explored. 
%A total of four object trackers, MedianFlow~\cite{kalal2010forward}, KCF~\cite{henriques2014high}, CSRT~\cite{lukezic2017discriminative}\textit{,} and OpticalFlow~\cite{kale2015moving}, are utilized and explored. -- SC: see comma that I inserted before the final and.

% Jay 10/29 Added resizing factor, to avoid confusion with the detector knob having the same name. Not sure if it is good enough, may be changed.
\noindent \textbf{Resizing factor - Input Image Resolution for Object Tracker}:
The resizing factor for input image of the object tracker is changed here such that a larger image requires the tracker to process through a larger number of pixels, with a reduction in efficiency. 

\noindent \textbf{Confidence Threshold of the Object Detector Backbone}:
The confidence threshold of the object detector backbone is closely related to the performance of the object tracker. A typical efficient detector backbone, such as EfficientDet or SSD, has a pre-defined number of detected objects or outputs (\eg detections with the top 100 confidence scores for both EfficientDet and SSD) to increase the accuracy performance of the model. 
A spike in energy and latency overhead is encountered if the tracker tracks all the detected objects. For example, the latency for tracking a single object of an 1280 x 720 image takes about 6.5 msec on the Xavier AGX board. For tracking 100 images, this value becomes 344.0 msec, which is more than 50 times degraded (higher) latency. Although the instantaneous power measurements are similar for both cases, since the energy consumption is accumulated power over time, the impact on latency affects energy consumption as well.
%
% \somali{can you give an example or two here to show the numbers especially because we say "huge".}
%JL110521: Did a quick experiment to add in some numbers.
Thus, we set a tunable threshold to control the number of objects to track.

\noindent \textbf{Detector Interval}:
\name leverages the usage frequency for the object detector backbone and the object tracker in the GoF. Every first frame in the GoF is passed through the object detector backbone to provide calibrated detection results for the object tracker, and the rest of the frames is passed through the object tracker. We define the number of frames in the GoF as the detector interval, indicating how often the detector should be run. For example, if the detector interval is 1, we run the object detector on all frames. In contrast, if the detector interval is 8, we run the object detector every 8 frames, and the rest with the object tracker.

\subsection{Efficient Multi-Branch Object Detection Kernel}\label{sec:multi_branch}

We propose our multi-branch object detection kernel as the combination of all possible combinations of the efficient methods or knobs listed in Sec.~\ref{sec:efficiency_knobs}. Particularly, an execution branch corresponds to a collection of choices on each efficiency knob and each branch can independently finish the task. Rigorously, an execution branch $b$ is denoted in a tuple form, 
\begin{equation}
    b = (d, rd, np, nm, t, rf, ct, i)
\end{equation}
where $d$ is the object detector backbone, $rd$ is the input resolution of the detector, $np$ is the number of proposals in the detector, $nm$ is the number of feature maps in the detector, $t$ is the object tracker, $rf$ is the resizing factor of the input image for object tracker, $ct$ is the confidence threshold to track, and $i$ is the detector interval. Thus, each execution branch is an instantiation in the high-dimensional configuration space, with a certain accuracy $a(b)$, energy consumption $e(b)$, and execution time (latency) $l(b)$. %However, due to the constraint of these methods, we cannot have the fully orthogonal set. 
However, these execution branch choices are not fully independent.
For example, if we choose EfficientDet as the object detector, we cannot use the ``Number of Proposals'', which is not applicable to EfficientDet. We further discuss the implementation details in Sec.~\ref{sec:implementation_multi_branch}. This notion of finding optimal configurations in a large configuration space with dependencies among different parameters has been tackled in other contexts, such as for distributed database tuning~\cite{mahgoub2020optimuscloud}.

% Jay 10/23 : Update caption.
\begin{figure}[t]
    \centering
    \begin{minipage}[c]{0.55\textwidth}
        \includegraphics[width=1\columnwidth]{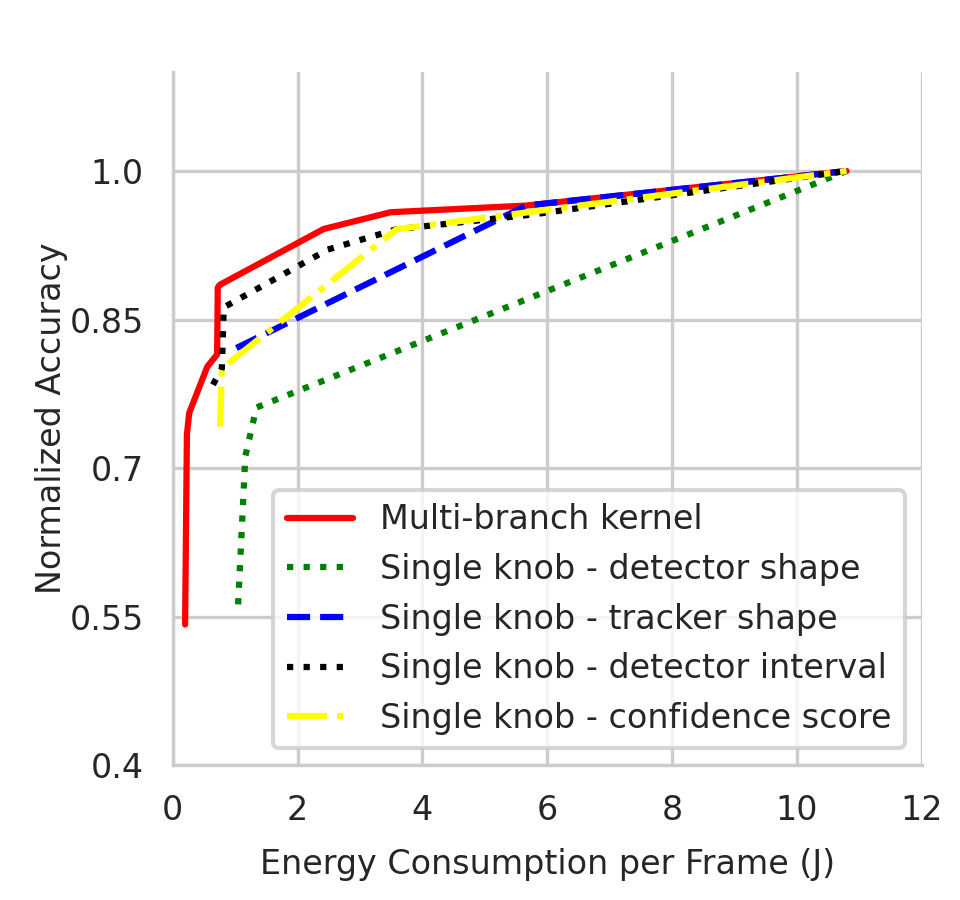}
        \caption{Accuracy of \name given multiple efficiency knobs versus single efficiency knob. The overall accuracy versus energy (same trend with accuracy vs. latency) tradeoff is always better using multiple knobs combined, compared to only using a single knob, validating the effectiveness of multiple knobs stacked together.}\label{fig:multi_branch}
    \end{minipage}
    \hfill
    \begin{minipage}[c]{0.4\textwidth}
        \includegraphics[width=1\columnwidth]{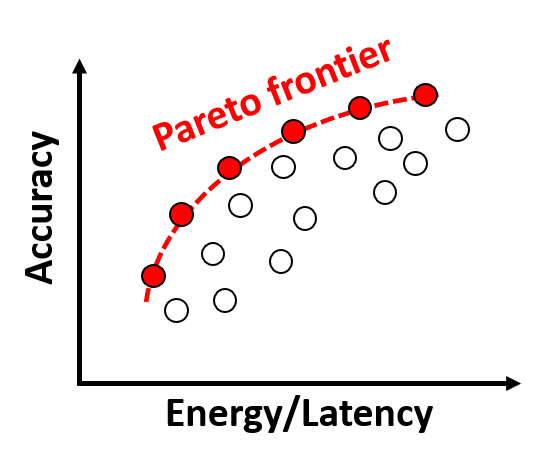}
        \caption{A Pareto frontier for the accuracy-energy/latency tradeoff for a particular model, by varying its efficiency knobs. The Pareto frontier has to be considered separately for each embedded device.}\label{fig:pareto}
        % \ran{The text must refer to this figure. Where is it?}.
        % % Jay 11/1 : edited the caption, no considering of contention level
        % % Jay 11/1 : Addressed, and referenced in section 3.3
        % % PC (11/02/2021): Updated figure
        % \somali{this should conceptually have energy consumption or latency in the X-axis}
        % \somali{Make sure you elaborate the captions of all figures and all figures are cross-referenced in the text.}
    \end{minipage}
\end{figure}

The advantage of multiple efficiency knobs over one knob is that their combination achieves a better Pareto optimal accuracy frontier for any efficiency requirement. In Fig.~\ref{fig:multi_branch}, we study the accuracy of \name along with each efficiency knob given a certain energy constraint compared to that of \name using multiple knobs. The results have shown that using multiple efficiency knobs gives a wider spectrum of accuracy vs. energy tradeoff and much higher accuracy at any energy level.

\subsection{Scheduler}\label{sec:secheduler}

% Ran (10/17/2021): enhance option: add the GPU memory as another user requirement
We design the scheduler in \name to select the most accurate branch $b_{opt}$ to satisfy users' requirements both with respect to the execution time (a budget of $l_0$ per video frame) and the energy (a budget of $e_0$ per frame), at runtime. 
% Jay 11/1 : added fig 3 reference here.
Fig.~\ref{fig:pareto} conceptually shows the selection of the scheduler, where among all possible execution branches, the scheduler picks the Pareto frontier performance branches that achieves highest performance in both latency or energy vs. accuracy.

Rigorously, the scheduler solves the following optimization problem: 
\begin{equation}
    \argmax_{b \in \mathcal{B}} a(b)~s.t.~e(b) \leq e_0,~l(b) \leq l_0
    \label{eq:scheduler}
\end{equation}
where $\mathcal{B}$ is the set of all possible branches in the multi-branch object detection kernel. $a(b)$, $e(b)$, and $l(b)$ are the accuracy, energy efficiency, and the latency of branch $b$, respectively. %Especially, 
The energy consumption has not been explored in existing studies on embedded object detectors. 
%
% Ran (10/17/2021): enhance option: add GPU memory as another user requirements
% GPU memory plays an important role when running video object detection models on embedded devices. Utilizing GPU for running deep neural networks comes with a big benefit in inference time, and sufficient GPU memory is required to execute a given object detection model.

% Jay 11/1 Minor modification. Changed the order to energy - latency - accuracy
As efficiency is one of the main drivers for the design of \name, the scheduler should also be lightweight, making immediate decisions as video frames arrive in the streaming style. To solve this problem, we model the energy consumption, latency, and accuracy of each execution branch in a data-driven manner. Particularly, we collect energy, latency, and the accuracy profile of each branch offline. Then, we train the energy, latency, and accuracy prediction models. We then use these models during the online phase so as to finish the task of the scheduler. When taking the choices of embedded devices, their power modes, and resource contention into consideration, these models are more complex than the simplified form in Eq.~\ref{eq:scheduler}, and our detailed design follows next.

\noindent \textbf{Energy Prediction Model}:
% \ran{Consider to move the energy prediction model above the latency one.}
% Jay 11/1 Addressed. Changed the order to energy - latency - accuracy
% Modified the energy prediction model section.
The energy consumption $e(b)$ of an execution branch is measured by calculating the average energy consumption of processing a single frame for each branch.
% is affected by many factors. For example, due to different computation capabilities of embedded boards, the energy consumption on each board is different. Also, the power mode of the device and the resource contention also affect the runtime energy consumption of an execution branch. 
We first profile the energy consumption on sample videos instead of the entire dataset, and measure the overall energy consumption of each execution branch. This is because the overall energy consumption of each execution branch is consistent across video frames and does not require such large amount of profiling data. 
% Jay 11/1 Maybe redundant information showing our weakness?
Since the exact energy consumption of a specific process on the embedded devices could not be measured, we use the overall energy consumption of the board as our metric.
We use the following equation, where $N$ represents the number of frames within the video, $p$ represents the instantaneous power measured at every 1 second interval, and $t$ represents the overall time of inference.

% \ran{Please explcitly write down the equation of the linear regression if it is similar to Eq.~\ref{eq:latency_profiling}.}
% Jay 11/1 Addressed. And double checked to find out that I actually manually profiled all the branches, since the tegrastats provides only the total energy, and we cannot separate the energy consumption of a specific process. Not sure if this is significant enough to be kept as a equation here.
\begin{equation}
    % e(b) = \frac{\Sigma{}}{n}
    e(b) = \frac{p*t}{N}
    \label{eq:energy_profiling}
\end{equation}

\noindent \textbf{Latency Prediction Model}:
The latency $l(b)$ of an execution branch is affected by many factors. For example, due to different computation capabilities of embedded boards, the latency on each board is different. Also, the power mode of the device and the resource contention also affect the runtime latency of an execution branch. To minimize the profiling cost, we use the following two techniques. \textit{First}, we profile the latency on sample videos instead of on the entire dataset. This is because the latency of each execution branch is consistent across video frames and does not require such large amount of profiling data. \textit{Second}, we decouple the profiling on the object detector and the object tracker. This allows us to profile all object detector branches and all object tracker branches, separately, and we use the following equation to calculate the overall latency due to the ``tracking-by-detection'' design,

\begin{equation}
    l(b) = \frac{l_{detector}(b) + (i-1) * l_{tracker}(b)}{i}
    \label{eq:latency_profiling}
\end{equation}

\noindent \textbf{Accuracy Prediction Model}:
The accuracy $a(b)$ of an execution branch is profiled in the offline training dataset and looked up in the online phase. The intuition is that the accuracy of each branch stays the same in the online phase since both the offline training dataset and the online test dataset follow the independent and identical distribution.
Considering the accuracy is meaningful given a large enough dataset and the number of execution branches is large, the cost of offline profiling is significant. Thus, we use the three following techniques to speed up the profiling. 
\textit{First}, we prune out the inferior branches in terms of accuracy and efficiency, and only use efficient yet effective models for the final design. For example, only SSD and EfficientDet are considered as the choices of object detectors. 
\textit{Second}, we use the high-end servers to profile the accuracy of each branch since our multi-branch execution kernel produces deterministic and consistent results between servers and embedded devices. \textit{Finally}, our profiling leverages the fact that the branches with same configurations except for detector internal $i$ can reuse the object detection results on the frames where object detector runs. We first profile the accuracy of all execution branches with $i=1$ (object detector only), save the detection results, and then profile the accuracy of other execution branches and reuse the saved detection results.

%\ran{Can we use an energy efficiency requirement here?}
%\jay{1. added energy and latency behind efficinecy requirement. 2. changed to  the low overhead of the branch prediction models}
% Ran: Thanks!
%
% Conclusion after introducing the prediction models.
To match stringent users' efficiency requirements --- energy or latency --- of inference at real-time (\eg 30 or 50 FPS) on embedded devices, the low overhead of the branch prediction models must be prioritized. Our implementation of light-weight prediction models comes with the benefit of low overhead. We empirically find that the overall latency overhead of our scheduler is less then 1 msec on a Jetson AGX Xavier board, which is marginal compared to the typical real-time frame rate of 30 FPS.

\subsection{Contention Generator}\label{sec:CG}

\begin{figure}[t]
    \centering
    \begin{minipage}[c]{0.3\columnwidth}
        \includegraphics[width=1\textwidth]{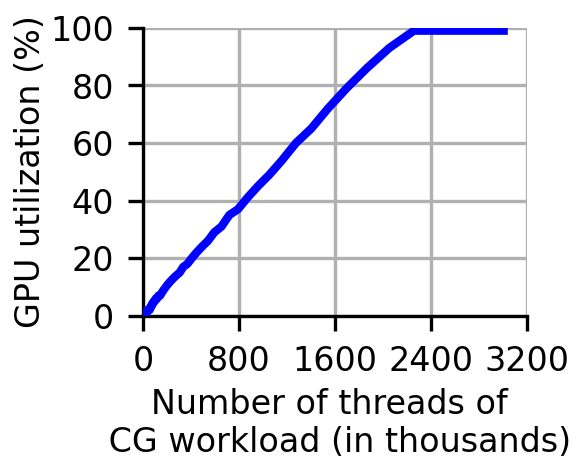}
        \centerline{\small (a) AGX Xavier}
    \end{minipage}
    \begin{minipage}[c]{0.3\columnwidth}
        \includegraphics[width=1\textwidth]{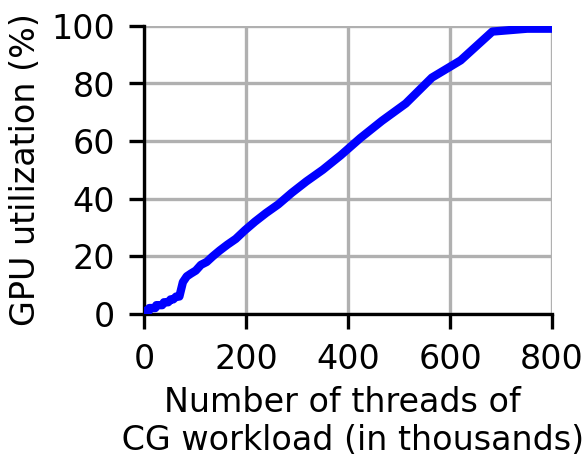}
        \centerline{\small (b) Xavier NX}
    \end{minipage}
    \begin{minipage}[c]{0.3\columnwidth}
        \includegraphics[width=1\textwidth]{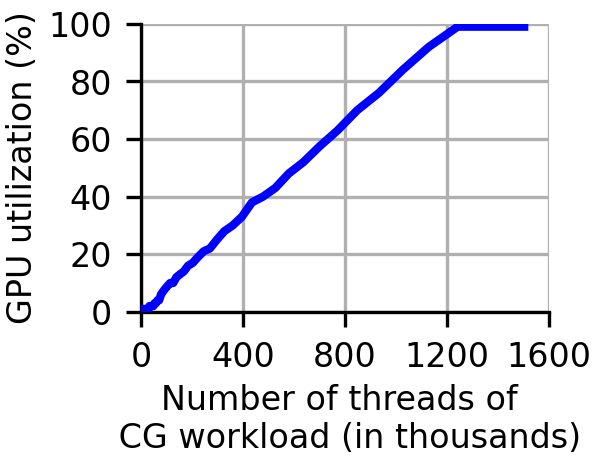}
        \centerline{\small (c) TX2}
    \end{minipage}
    \caption{Calibration of the contention generator on each embedded device.}\label{fig:calibration}
\end{figure}

To understand the performance of object detection models under resource competition, we design and implement a synthetic contention generator (CG) to create a tunable amount of GPU resource contention to the object detection models. The CG is a stand-in for the general background and concurrent workloads executed on the device, which consume GPU resources. With the CG, we are able to profile the performance of all models under different resource contention scenarios. 

The CG occupies designated levels of resource on the GPU module of the embedded device by a percentage of the maximum capacity. To achieve this goal, the CG on the GPU performs \textit{add operation} with a CUDA kernel function. By changing the number of threads of the CG workload, we control the number of GPU cores that are kept busy per second. Thus, we are able to occupy different amounts of GPU resources and we call the amount of GPU resource that the CG occupies the \textit{GPU contention level}. We choose 12 GPU contention levels [0\%, 1\%, 10\%, 20\%, 30\%, 40\%, 50\%, 60\%, 70\%, 80\%, 90\%, 99\%] since they cover the usable GPU resource space as comprehensively as possible with a reasonable number of levels.

Since each embedded device comes with a different GPU architecture, number of cores, and computation capabilities, a CG workload with a certain number of threads results in different GPU contention levels on different boards. We calibrate the CG on each embedded device (namely, NVIDIA Jetson AGX Xavier, Xavier NX, and TX2 boards), to produce a consistent level of contention. We show our calibration experiment results of the CG on AGX Xavier, Xavier NX, and TX2 in Fig.~\ref{fig:calibration}. As can be seen, with the increasing number of threads of the CG workload, the GPU contention level (GPU utilization) keeps increasing and finally saturates at 99\% as the number of threads is large enough. One key observation is that the relationship between the number of threads of CG workload and the GPU contention level is linear, which makes our CG a small linear control system. 

\section{Implementation}\label{sec:implementations}

In this section, we mainly describe the implementation details of our multi-branch object detection kernel (Sec.~\ref{sec:implementation_multi_branch}), training details of object detectors (Sec.~\ref{sec:training_efficientdet}), and the embedded devices we used for evaluation (Sec.~\ref{sec:embedded}).

\subsection{Efficient Multi-Branch Object Detection Kernel}\label{sec:implementation_multi_branch}

We implement our efficient multi-branch object detection kernel with the following software stacks: CUDA 10.2.89 and CuDNN 8.0.0.180 as the base libraries for GPU-accelerated DNN primitives, and TensorFlow 2.4.0 with Python 3.6, for developing the overall framework. Also, given the multiple efficiency knobs we have implemented, there exists a vast number of execution branch combinations. Thus, we propose the following implementation techniques to reduce both offline and online costs to a practically feasible level.

\noindent \textbf{Pruning the Efficiency Knobs}:
First, we only consider EfficientDet and SSD as the object detector backbone in \name, as they show superior performance in energy, latency, and accuracy, compared to FRCNN and YOLO.
% SB (11/7/21): Not clear what superior accuracy vs. efficiency means.
% Jay 11/9 : Addressed. Slightly reworded to be more specific.
Also these two object detectors are optimized for performance on mobile devices. While \name only incorporates EfficientDet and SSD due to its efficient design, we still include FRCNN and YOLO with other efficiency knobs, and use them as baselines for evaluation.
Next, we narrow down the efficiency knobs to be applied to each object detector backbone. The ``number of proposals'' knob is exclusive to FRCNN, so it is not included in \name. For EfficientDet D0 and D3, the models come with the built-in input image resolution that is hard-coded, matching their scaling factor and the design of the feature pyramid architecture (what they refer to as the Bi-FPN). Therefore, the input image resolution knob is implemented only for SSD. 
In addition, we have tested the number of feature map knobs that are exclusive to SSD, and check the accuracy and latency of all possible combinations on the ILSVRC 2015 VID dataset. 
% Jay 11/1 : Added a bit more explanation.
Our intuition of using a subset of the feature pyramid layers in MnasFPN is that reducing the number of layers used to compute the features will result in benefit in latency with a moderate tradeoff of accuracy.
However, due to the explicit design of the MnasFPN generated by the NAS technique, each feature level had a significant drop in the accuracy, while removing feature levels afforded limited latency reduction benefit.
%SC110121: Explain how MnasFPN comes into the picture. 
% Jay 11/1 : The overall methodology for MnasFPN is explained in section 3.1.1

% Jay 11/1 : Commented below, since it is not significant, and I think it does not blend in well with our current storyline.
% -------------------------------------------------------------------------
% % Jay (10/18/2021): Maybe the below part could be optional. But put it in just in case
% One interesting result to note is that removing the last feature pyramid level when using the smallest input image resolution of 192, the accuracy is the same with a small benefit in latency. 
% %don't say gain in latency as in the word gain because it makes it look like a good thing, instead say increase or decrease -- you mean increase in latency above?
% % Jay 11/1 : replaced with benefit
% This is because the last feature pyramid level is responsible for detecting small-scaled objects, and the small input image resolution results in the loss of most of the information to be processed. However, utilizing this tuning knob just for one combination has introduced more overhead compared to the gain from it, so we have not included in our final implementation.
% \somali{can you clarify the writing above, the logic does not flow well. The tuning knob you are talking about is the number of layers of the feature pyramid.}
% -------------------------------------------------------------------------

\noindent \textbf{Defining a Set of Choices for Each Knob}:
Most efficiency knobs, \eg input resolution and detector interval, can support any integer choices. However, due to cost of the offline profiling and the online scheduling, we define a discrete set of choices for each knob as follows. 
The SSD and EfficientDet D0 and D3 are selected as the object detector backbones. Input image resolution for the detector is implemented for the SSD, and we use the shape of [192, 256, 320] as our pre-defined set of choices. The acceptable image resolution depends on the network architecture.
% , \ie the resolution changes in convolution layers. 
MedianFlow tracker is selected for its light-weight design
%of being much faster
compared to the other trackers we have explored, while maintaining comparable accuracy. 
%SC110121: So we hand-pick MedianFlow tracker? And eliminate the other trackers from our design? Rather than \name automatically deciding which tracker to use?
% Jay 11/1 : Yes, to maintain real-time latency on embedded devices, most other trackers are quite heavy. Also the accuracy profiling showed best performance for MedianFlow as well.
The input resolution for the tracker is [100\%, 50\%, 25\%] of the original resolution on height and width dimensions. For detector interval, we have [1, 2, 4, 8, 20, 100], and finally, we use confidence thresholds of [0.15, 0.30] to post-process the detection outputs from the object detector. The confidence threshold removes objects with lower confidence scores, and controls the number of objects that need to be tracked by the object tracker.
%
% Jay 10/27 relocated prof Chaterji's comments. Need to be addressed after relocation of above content.
% Ran (10/30/2021): addressed.
% \somali{We only use D0 and D3. Mention that and argue for our choice.}
% \ran{Check the first paragraph of ``Pruning the efficiency knobs''. We have clearly explained.}
% \somali{This description of EfficientDet is relevant for Sec 5. So I think this needs to come toward the beginning of that section.}
% \ran{They are implementation details and I think it is the right place here.}

% For internal purposes.
% For SSD : 3 detector only + 5 si [2, 4, 8, 20, 100] x 2 conf [0.15, 0.3] x 3 ds [1, 2, 4] x 3 shape [192, 256, 320] = 93 branches
% For EfficientDet : 2 detector only + 5 si x 2 conf x 3ds x 2 model [d0, d3] = 62 branches
% A total of 155 branches
% Jay 10/28: Reworded.
% Ran (10/30/2021): Commented "a total of 93 execution branches in the SSD+ multi-branch object detector kernel, and 31 execution branches for each EfficientDet D0 and D3 multi-branch object detector kernel, having", because we have not introduced SSD+ or D0+. No need to count in details.
Considering all the pre-defined efficiency knobs above, we have a total 155 execution branches for \name.
Having hundreds of branches in a multi-branch object detection kernel does not mean we have to store and load that many copies of branches in the disk or the memory. 
%SC110121: Do you mean we do not have to store and load that many different models?
% Jay 11/1 : Yes, inside a single Tensorflow session, with a single model weight file, we can do all the switching of branches. The above sentence refers to maintaining a model weight file for each execution branch.
``Input resolution of the object detector backbone'', ``number of proposals in the detector backbone'', ``number of feature maps in the detector backbone'', ``resizing factor of the object tracker'', ``confidence threshold to track'', and ``detector interval'' can all be implemented as a control parameter with just one copy of the model.
% stored in the disk and loaded in the memory. 
As for the choice of the object detector backbone and object tracker, we lower the switching cost among the execution branches by merging the static computation graphs of all object detector backbones. The benefit of merging the static graph is that the each object detection backbone is only loaded into the GPU memory when it is executed, which saves resources for detector backbones not being used. In addition, the executed object detector backbones are preserved in a cached state, and \name is able to switch among loaded backbones with minimal overhead without requiring the initialization of the detector backbone.

\subsection{Training Efficient Object Detectors}\label{sec:training_efficientdet}

EfficientDet D0, D3, and SSD object detectors do not come with publicly available pretrained weights for the ILSVRC 2015 VID datasets~\cite{ILSVRC15}. Following the widely adopted training protocols in the latest video object detection solutions that we have selected as baselines for evaluation~\cite{chen2020memory, sabater2020robust, wu2019sequence}, we train each object detector on a combined dataset of ILSVRC VID training and DET training datasets~\cite{ILSVRC15}. The ILSVRC 2015 VID training dataset consists of 3,862 videos with 30 object classes. 
From each video, 15 video frames, whose timestamps are evenly spaced, are selected. As an addition to the ILSVRC 2015 VID training dataset, images containing the 30 overlapping classes with the VID dataset are selected from the DET training dataset. Finally, a total of 111,473 video frames or images, 57,834 from the VID dataset, and 53,639 from the DET dataset, are selected for the training process. We train EfficientDet D0 and D3 from the COCO pre-trained weights that come from the official repository and follow the default training settings, 
% except that we use a different set of class labels and number of classes, 
while freezing the backbone part (EfficientNet) during the finetuning.
%SC110121: Why "a different set of class labels and number of classes"?
% Jay 11/1 : I think the original writing meant to say the difference between COCO and ILSVRC, but just commented that part.
To speed up, we use 1\% of ILSVRC 2015 VID training dataset (11,768 video frames) for the first 100 epochs of D0 and for the first 50 epochs of D3, and then use the aforementioned VID and DET dataset for the remaining 10 epochs. In addition, the SSD model combined with MobileNetV2 and MnasFPN also comes with the COCO pre-trained weights from the official repository (not our evaluation dataset ILSVRC 2015 VID). We follow most of the default training settings from the repository, except for using a batch size of 48 and a learning rate of 0.004. The model is trained with the aforementioned VID and DET dataset, and is trained up to 180 epochs.
% Ran (10/26/2021): commented ``and checkpoint at 163 epochs is selected as the best model''.
%SC110121: So convergence is reached by 180 epochs?
% Jay 11/1 : Yes. Accuracy dropped beyond that point.

\subsection{Embedded Devices}\label{sec:embedded}

% \ran{Controdictory writing on power mode on AGX. The first part says mode 0 and second part says 0 and 2}
% PC (10/28/2021): Commented out the description on power mode.
% \ran{Re-open the issue: we fix the power modes in our experiments for reproducibility---mode 0 on AGX Xavier. We used two power modes, right? Why do we say we fix the power mode here?}
% PC (11/01/2021): Addressed
We evaluate \name and baseline models on NVIDIA Jetson AGX Xavier~\cite{agxxavier}, Jetson Xavier NX~\cite{xaviernx}, and Jetson TX2~\cite{tx2}. Each device has different CPU, GPU, and memory capacities, and the relationship between their computational capacities is Jetson AGX Xavier $>$ Jetson Xavier NX $>$ Jetson TX2. Table~\ref{tab:jetson} gives the hardware specifications of these devices. Each board has different numbers of power mode levels shown in Table~\ref{tab:agx-power-modes} for Jetson Xavier AGX, and Table~\ref{tab:nx-power-modes} for Jetson NX Xavier. We picked some power modes in our experiments to better understand the power modes of Jetson devices --- mode 0 and mode 2 on AGX Xavier, mode 0, 2, and 4 on Xavier NX, and mode 0 on TX2~\footnote{The default modes are: mode 7 on AGX Xavier, mode 3 on Xavier NX, and mode 3 on TX2.}. These levels can impact the performance of an object detection system by altering the maximum power budget, maximum frequency for CPU, GPU, and the deep learning accelerator, and number of online CPU cores. 
%
% We also implement the power modes as an efficient knob in our experiments for handling the trade off between energy and latency efficiency. We evaluate the power mode tuning knob on AGX Xavier, and select two power modes --- 0 and 2 in Table~\ref{tab:agx-power-modes} to represent the strongest and weakest power mode without a reboot\footnote{Power mode 1 is the weakest power mode on Jetson AGX Xavier, and however, switching to power mode 1 requires a complete reboot and reconfiguration of the board. Thus, we select the second weakest power mode, power mode 2, to be our weakest selection.}. 
% %
% % Ran: what does ``static and dynamic power consumption'' mean?
% \ran{``fixing the frequency of the modules at their max frequencies'' is contradictory to varying power mode on AGX. Should we say sticking to a certain power mode?}
% PC (11/01/2021): Changed to "max frequencies under the corresponding power modes"
Furthermore, these devices have a native Dynamic Voltage and Frequency Scaling (DVFS) functionality on the CPU and GPU, enabled by default. DVFS provides a way to reduce static and dynamic power consumption of the embedded boards on the fly by scaling up or down the voltage and frequency based on the targeted performance of the application~\cite{jetson_guide}. We find that the default DVFS functionality can cause inconsistency in our evaluation results because of the changing CPU, GPU, and memory frequencies. Therefore, we disable DVFS by fixing the frequency of the modules at their max frequencies under the corresponding power mode. We empirically determine that this step is crucial to reproducibility of results.

\begin{table*}[htbp]
    \centering
    \small
    \begin{tabularx}{\textwidth}{|>{\hsize=.55\hsize\centering\arraybackslash}X
    |>{\hsize=1.15\hsize\centering\arraybackslash}X
    |>{\hsize=1.15\hsize\centering\arraybackslash}X
    |>{\hsize=1.15\hsize\centering\arraybackslash}X|}
    \hline
    \textbf{Models} & \textbf{Jetson AGX Xavier} & \textbf{Jetson Xavier NX} & \textbf{Jetson TX2}\\ 
    \hline
    \hline
     CPU & 8-core NVIDIA Carmel Armv8.2 64-bit CPU 8MB L2 + 4MB L3 with max frequency at 2265MHz & 6-core NVIDIA Carmel ARMv8.2 64-bit CPU 6MB L2 + 4MB L3 with max frequency at 1900MHz & Dual-Core NVIDIA Denver 2 64-Bit CPU and Quad-Core ARM Cortex-A57 MPCore processor with max frequency at 2000MHz\\
    \hline
    GPU & 512-core NVIDIA Volta GPU with 64 Tensor Cores with max frequency at 1377MHz & 384-core NVIDIA Volta GPU with 48 Tensor Cores with max frequency at 1100MHz & 256-core NVIDIA Pascal GPU with max frequency at 1300MHz \\
    \hline
    Memory & 32 GB 256-bit LPDDR4x 136.5GB/s & 8 GB 128-bit LPDDR4x 51.2GB/s & 8 GB 128-bit LPDDR4 59.7GB/s \\
    \hline
    Storage & 32 GB eMMC 5.1 & 16 GB eMMC 5.1 & 32 GB eMMC 5.1 \\
    \hline
    Power & 10W/15W/30W & 10W/15W & 7.5W/15W \\
    \hline
    DL Accelerator & 2x NVDLA Engines & 2x NVDLA Engines & - \\
    \hline
    AI Performance & 16 TFLOPS & 10.5 TFLOPS & 1.33 TFLOPS \\
    %YL112321: TOPS are Nvidia's claim on tensor operations. Here we should also list and compare the actual FLOPS.
    %JL122321: Changed TOPS to TFLOPS by dividing by 2
    % https://developer.nvidia.com/blog/nvidia-jetson-agx-xavier-32-teraops-ai-robotics/
    % Check VOLTA GPU section.
    \hline
    Price & \$699 & \$399 & \$399 \\
    \hline
    \end{tabularx} 
    \caption{Specifications for NVIDIA Jetson devices: AGX Xavier, Xavier NX, and TX2.}\label{tab:jetson}
    % \ran{Switching the columns of AGX and TX2. Change the ordering in the title.}
    % PC (10/29/2021): Done
\end{table*}

\begin{table}[ht!]
    \centering
    \small
    \begin{tabularx}{\columnwidth}{|>{\hsize=0.2\hsize\centering\arraybackslash}X
    |>{\hsize=.1\hsize\centering\arraybackslash}X
    |>{\hsize=.1\hsize\centering\arraybackslash}X
    |>{\hsize=.1\hsize\centering\arraybackslash}X
    |>{\hsize=.1\hsize\centering\arraybackslash}X
    |>{\hsize=.1\hsize\centering\arraybackslash}X
    |>{\hsize=.1\hsize\centering\arraybackslash}X
    |>{\hsize=.1\hsize\centering\arraybackslash}X
    |>{\hsize=.1\hsize\centering\arraybackslash}X|}
        \hline
        \textbf{Key Parameters} & \textbf{Power Mode 0} & \textbf{Power Mode 1} & \textbf{Power Mode 2} & \textbf{Power Mode 3} & \textbf{Power Mode 4} & \textbf{Power Mode 5} & \textbf{Power Mode 6} & \textbf{Power Mode 7}\\ 
        \hline
        \hline
        Power Budget & N/A & 10W & 15W & 30W & 30W & 30W & 30W & 15W\\
        \hline
        Online CPU Cores & 8 & 2 & 4 & 8 & 6 & 4 & 2 & 4\\
        \hline
        Maximal CPU Frequency (MHz) & 2265.6 & 1200 & 1200 & 1200 & 1450 & 1780 & 2100 & 2188\\
        \hline
        Maximal GPU Frequency (MHz) & 1377 & 520 & 670 & 900 & 900 & 900 & 900 & 670\\
        \hline
        %DL Accelerator cores & 2 & 2 & 2 & 2 & 2 \\
        %\hline
        Maximal DL Accelerator Frequency (MHz) & 1395.2 & 550 & 750 & 1050 & 1050 & 1050 & 1050 & 115.2\\
        \hline
        %Maximal Memory Frequency (MHz) & 1600 & 1600 & 1600 & 1600 & 1600 \\
        %\hline
    \end{tabularx}
    \caption{Different power modes on the Jetson AGX Xavier board. Each power mode comes with a different maximum power budget, number of online CPU cores, CPU, GPU, and DL accelerator frequency, which gives flexibility toward selecting the energy efficiency.}\label{tab:agx-power-modes}
\end{table}

% Jay 11/3 : Relocated.

\begin{table}[ht!]
    \centering
    \small
    \begin{tabularx}{\columnwidth}{|>{\hsize=0.4\hsize\centering\arraybackslash}X
    |>{\hsize=.1\hsize\centering\arraybackslash}X
    |>{\hsize=.1\hsize\centering\arraybackslash}X
    |>{\hsize=.1\hsize\centering\arraybackslash}X
    |>{\hsize=.1\hsize\centering\arraybackslash}X
    |>{\hsize=.1\hsize\centering\arraybackslash}X|}
        \hline
        \textbf{Key Parameters} & \textbf{Power Mode 0} & \textbf{Power Mode 1} & \textbf{Power Mode 2} & \textbf{Power Mode 3} & \textbf{Power Mode 4}\\ 
        \hline
        \hline
        Power Budget & 15W & 15W & 15W & 10W & 10W \\
        \hline
        Online CPU Cores & 2 & 4 & 6 & 2 & 4 \\
        \hline
        Maximal CPU Frequency (MHz) & 1900 & 1400 & 1400 & 1500 & 1200 \\
        \hline
        Maximal GPU Frequency (MHz) & 1100 & 1100 & 1100 & 800 & 800 \\
        \hline
        %DL Accelerator cores & 2 & 2 & 2 & 2 & 2 \\
        %\hline
        Maximal DL Accelerator Frequency (MHz) & 1100 & 1100 & 1100 & 900 & 900 \\
        \hline
        %Maximal Memory Frequency (MHz) & 1600 & 1600 & 1600 & 1600 & 1600 \\
        %\hline
    \end{tabularx}
    \caption{Different Power Modes for Jetson Xavier NX.}\label{tab:nx-power-modes}
\end{table}
% \somali{This table can be removed since we do not give energy experiments on the Xavier NX.}
% Ran (10/20/2021): we do have. So keep it.
\section{Evaluation}\label{sec:evaluation}

% Left TODOs for Jay 10/29: 
% 1. Add more baselines(e.g. AdaScale in figure 5.)
% 2. Further polish figures.

% Jay : Things to look during a pass.
% Terminology
% Use efficiency knobs to address the tuning knobs.
% Use object detector / object detector backbone / object detector baselines to address detector models.
% Use multi-branch object detection kernel to mention the object detector backbone coupled with our efficiency knobs.
% Try not to use models.
% Remove the term 'benchmark' replace with 'evaluation'

To evaluate \name, we first introduce all multi-branch object detection kernels of \name and baselines in Sec.~\ref{sec:baselines}, and evaluation dataset and metrics in Sec.~\ref{sec:metrics_dataset}. 
% Jay 11/3 : Added Key takeaway summary in the beginning. Not sure if the format is appropriate, or it has the right level of details. Would very appreciate any feedback.
We then present our evaluation results in the following sections:
\begin{enumerate}
    % 1. Overall performance of \name + different energy or latency requirements with \name.
    \item First, we show the comparison of \name to different baselines. In addition, we evaluate \name under both energy consumption or latency requirements, and present the results in Sec.~\ref{sec:efficiency_requirements}. 
    % 2. Impact of power mode on energy vs. latency tradeoff
    \item Second, we rigorously investigate the impact of power mode on energy vs. latency tradeoff during the evaluation of \name with different user requirements, and show that power modes on embedded devices can be utilized to achieve further optimization on energy or latency performance.
    % 3. different devices, contention, multiple baselines for accuracy vs. latency
    \item Third, we evaluate \name and all baselines comprehensively on different embedded devices under various resource contention scenarios for accuracy, latency, and energy in Sec.~\ref{sec:accuracy_latency_benchmark}, present the impact of runtime environment on object detectors, and further show the effectiveness of \name's adaptive features.
    % 4. more in-depth analysis of energy.
    \item Finally, in Sec.~\ref{sec:energy-exp} we present a more in-depth energy consumption analysis on our multi-branch object detection kernels and baselines by utilizing more fine-grained power modes.
\end{enumerate}

% Jay 10/27 : Remove the term SOTA, and name them as baseline models
% Ran (10/27/2021): Enhance option 1 add AdaScale [MLSys]
% Ran (10/27/2021): Enhance option 2 add ApproxDet [SenSys'19]
% Ran (10/27/2021): Enhance option 3 add DorT [AAAI'19]
\subsection{\name Variants and Baselines}\label{sec:baselines}

% Jay 10/28 : updated to 9 solutions and 15 protocols.
% solutions : d0, d3, ssd, frcnn, yolo, fastadapt, repp, mega, selsa
% protocols
% Jay 10/28 : textit{D0+} & \textit{D0} & \textit{D3+} & \textit{D3} & \textit{SSD+} & \textit{SSD} & \textit{FRCNN+} & \textit{FRCNN} & \textit{YOLO+} & \textit{YOLO} & \textit{Fast Adapt} & \textit{REPP w\ YOLOv3} &  \textit{MEGA}  & \textit{SELSA 50} & \textit{SELSA 101}

We consider 9 baseline video object detection solutions and their variants, with a total of 15 protocols, for our evaluation. These models are selected using the following criteria:
\begin{enumerate}[leftmargin=1em]
    \item The code and model could be deployed on a Jetson TX2 board, and
    \item The model is open-sourced and can be replicated on the ImageNet Large Scale Visual Recognition Challenge (ILSVRC) 2015 VID dataset~\cite{ILSVRC15}.
    %SC102021: The second reason is a weak point of our paper. A reviewer may say take the real SOTA and train on your dataset if you need to. 
    % Ran (10/30/2021): change to ``replicable''.
    %SC -- good, wordsmithed a bit.
\end{enumerate}

The detailed explanation of each protocol is as follows.

\subsubsection{Adaptive Video Object Detection Models} \hfill\\

\noindent We implement and evaluate a total of 6 adaptive video object detection models. 

% Jay 11/2 : Chopped and merged with D0+ and D3+
% \ran{The Description of D0 and D3 is a little redundant since they are our model and we already introduce them in the implementation section. Consider to chop the duplicated sentences.}
% \noindent{\bf EfficientDet D0 and D3}
% EfficientDet~\cite{tan2020efficientdet} is an efficiency oriented object detector that uses scaling to generate multiple model families. There are eight models from D0 to D7, with different scaling factors that give different depth, width, and image resolution for the network. Our observation is that heavier model variants above D3 cannot run on the Jetson TX2 board due to insufficient memory. Thus, we select EfficientDet D0 and D3 as kernel candidates as the most light-weight and heavy-weight models among the executable variants. 

% \ran{What does ``part of the object detector backbone'' mean?}
% Jay 11/2 : reworded. originally meant, it is a kernel for \name.
% Ran: Thanks!

\noindent{\bf EfficientDet D0, D0+ and D3, D3+}: EfficientDet D0+ and D3+ are multi-branch object detection kernels, which is our improvement over EfficientDet D0 and D3 with efficiency knobs. We have four efficiency knobs for EfficientDet D0+ and D3+, as follows: 1) object tracker, 2) input image resolution for the object tracker, 3) confidence threshold to track, and 4) detector interval. 
%SC103021: editorial -- in American English, if there are more than 2 items, you put a comma before the final "and".

% Jay 11/2 : commented detailed explanation
% Coupled with the MedianFlow tracker, we use [100\%, 50\%, 25\%] for resizing the input image resolution of the object tracker, and [15\%, 30\%] confidence threshold, and the set of [1, 2, 4, 8, 20, 100] for the detector interval.

% \ran{SSD+ is not an object detector backbone. It represents a collection of execution branches that uses SSD as the object detector backbone.}
% Jay 11/2 : reworded and added the term multibranch object detection kernel.
% \ran{Relocate the implementation details to the implementation section.}
% Jay 11/2 : Addressed.
\noindent{\bf SSD and SSD+}: SSD+ is another multi-branch object detection kernel, that is SSD as the object detector backbone combined with our efficiency knobs. There are a total of five efficiency knobs for SSD+: 1) input image resolution for the object detector backbone, 2) object tracker, 3) input image resolution for the object tracker, 4) confidence threshold to track and 5) detector interval. 

% Jay 11/2 : Chopped off details for implementation.
% We use input image resolution of [192, 256, 320] for the object detector, and use a MedianFlow tracker. In addition, the specific set for the tracker-related efficiency knobs are as follows: [100\%, 50\%, 25\%] for the resizing factor, and [15\%, 30\%] confidence threshold, and the set of [1, 2, 4, 8, 20, 100] for the detector interval.

\noindent{\bf FastAdapt}: FastAdapt~\cite{lee2021benchmarking} is an adaptive framework that is able to adapt to different latency requirements. We use FastAdapt as one of the adaptive baselines for performance comparison. FastAdapt uses a single object detector backbone, FRCNN, and also incorporates an object tracker to speed up the average inference latency.

\noindent{\bf FRCNN and FRCNN+}: FRCNN+ is our improvement over FRCNN~\cite{ren2015faster} and
we added two efficiency knobs for FRCNN+: 1) image shape and 2) number of proposals. With different combinations of the image shape and number of proposals, there are 28 execution branches with the input image resolution in the following set [224, 320, 448, 576] and number of proposals in the following set [1, 3, 5, 10, 20, 50, 100]. We use the MedianFlow tracker while keeping a detector interval of 8 frames, which is a middle-of-the-range value.
% among our full set, to limit the search space and show the performance of our combination of efficiency knobs.
%SC103121: Didn't understand the part commented out above
% Jay 11/1 Was meant to describe the reason why we only used si of 8.

\noindent{\bf YOLO and YOLO+}: YOLO+ is our improvement over YOLO~\cite{redmon2016you, redmon2018yolov3}, combined with some of the efficiency knobs for limited adaptivity. We included one efficiency knob for YOLO, which is the shape of the input image, and also used it in tandem with the MedianFlow object tracker~\cite{kalal2010forward} for acceleration. There are a total of 12 execution branches determined by the input image resolution in the following set [224, 256, 288, 320, 352, 384, 416, 448, 480, 512, 544, 576]. Similar to FRCNN+, we also limit the detector interval to 8 frames for YOLO+.

\subsubsection{Accuracy-Optimized Video Object Detection Models}  \hfill\\

% Jay 10/28: TODO: Check later to add more details and citations.
\noindent Several latest video detection models are considered to provide baseline experiments of {\em non-adaptive} models. These models are optimized for accuracy.

\noindent{\bf REPP}: REPP~\cite{sabater2020robust} comes with a total of three model variants --- YOLOv3, SELSA, and FGFA. However, only their implementation over a YOLOv3 baseline is able to run on a TX2 board. We address this baseline as ``REPP with YOLOv3'' to avoid confusion with our own YOLO implementations (YOLO and YOLO+).

\noindent{\bf SELSA}: For SELSA~\cite{wu2019sequence}, we utilize ResNet-50 and ResNet-101, with the corresponding variants referred to as SELSA 50 and SELSA 101.

\noindent{\bf MEGA}: MEGA~\cite{chen2020memory} is provided with two different object detector backbones with different levels of features usage. MEGA utilizes a feature aggregation technique to further improve accuracy by capturing the content similarity for both neighboring frames and overall global frames within the video. However, this feature considers and processes multiple frames at runtime, thus cannot run on embedded devices due to lack of memory. 
Here, we only use the object detector backbone provided by the authors with ResNet-50 as the feature extractor and limit the use of feature aggregation.
% YL112321: Isn't this the same as Faster RCNN then?
% We call this model MEGA base.
% \ran{Why we call it base?}
% \somali{MEGA base does sound confusing, reword}
% Jay 11/2 Addressed. We called it base since we only used the detector backbone without any frame aggregation. Added more explanations, and just call it mega.

\subsection{Evaluation Dataset and Metrics}\label{sec:metrics_dataset}

\noindent \textbf{Dataset}: We use the ILSVRC 2015 VID validation dataset~\cite{ILSVRC15} as the evaluation dataset for the video object detection task---to classify and localize the objects in 30 classes, over 555 videos (176,126 frames in total). 
% \ran{Suggest to relocate to Section 5.2 or comment. The clalrification of the ``streaming'' setting should apply to all experiments.}
% Jay 11/1 : Relocated.
For adaptive video object detection baselines, we consider a streaming setting for inference with video frames fed one-by-one, and report the mean latency per video frame. For other methods, we use batch processing for inference with a batch size of 1 to feed the frames one-by-one.
% \somali{A reviewer may question how it is fair to compare latency or accuracy between streaming mode and batch mode.}
% \jay{Tried to address it by adding more details for batchsize 1 inference.}

\noindent \textbf{Accuracy} of all baselines is measured in terms of mean average precision (mAP), following the widely adapted evaluation protocol~\cite{lin2014microsoft} on the dataset. mAP is defined as the mean of APs for all classes. AP is the area under the precision-recall curve that measures both localization and classification accuracy by comparing detection bounding boxes against ground-truth boxes using a fixed Intersection-over-Union (IoU) threshold of 0.5.

\noindent \textbf{Energy consumption} is measured by the native API provided by the Jetson board {\em tegrastats}~\cite{tegrastats} utility from NVIDIA. {\em Tegrastats} provides the information of instantaneous power usage of the CPU and GPU modules, and we convert the power consumption to the overall energy consumption, and then divide it by the number of frames to calculate the average energy consumption per frame. Users can also specify the interval of the execution of {\em tegrastats}. 
We use a 1-second interval in our experiments. This choice collects accurate measurements at a fine time granularity while bounding the overhead of the measurement and corresponding impact on the performance during evaluation.

\noindent \textbf{Latency} of all baselines are measured as per-frame latency over the Group of Frame (GoF) during inference\footnote{Due to our tracking-by-detection technique in Sec.~\ref{sec:efficiency_knobs}, the latency of the first frame and remaining frames are uneven. Thus, we take the average over a GoF as the temporal latency.}, and further aggregated over GoFs of all videos on the dataset.

\subsection{Satisfying Various Efficiency Requirements} \label{sec:efficiency_requirements}

%%%%%%%%%%%%%%%%%%%%%%%%%%%%
%%%%%%%   Figure 5   %%%%%%%
%%%%%%%%%%%%%%%%%%%%%%%%%%%%
\begin{figure}[t]
    \centering
    \includegraphics[width=0.85\textwidth]{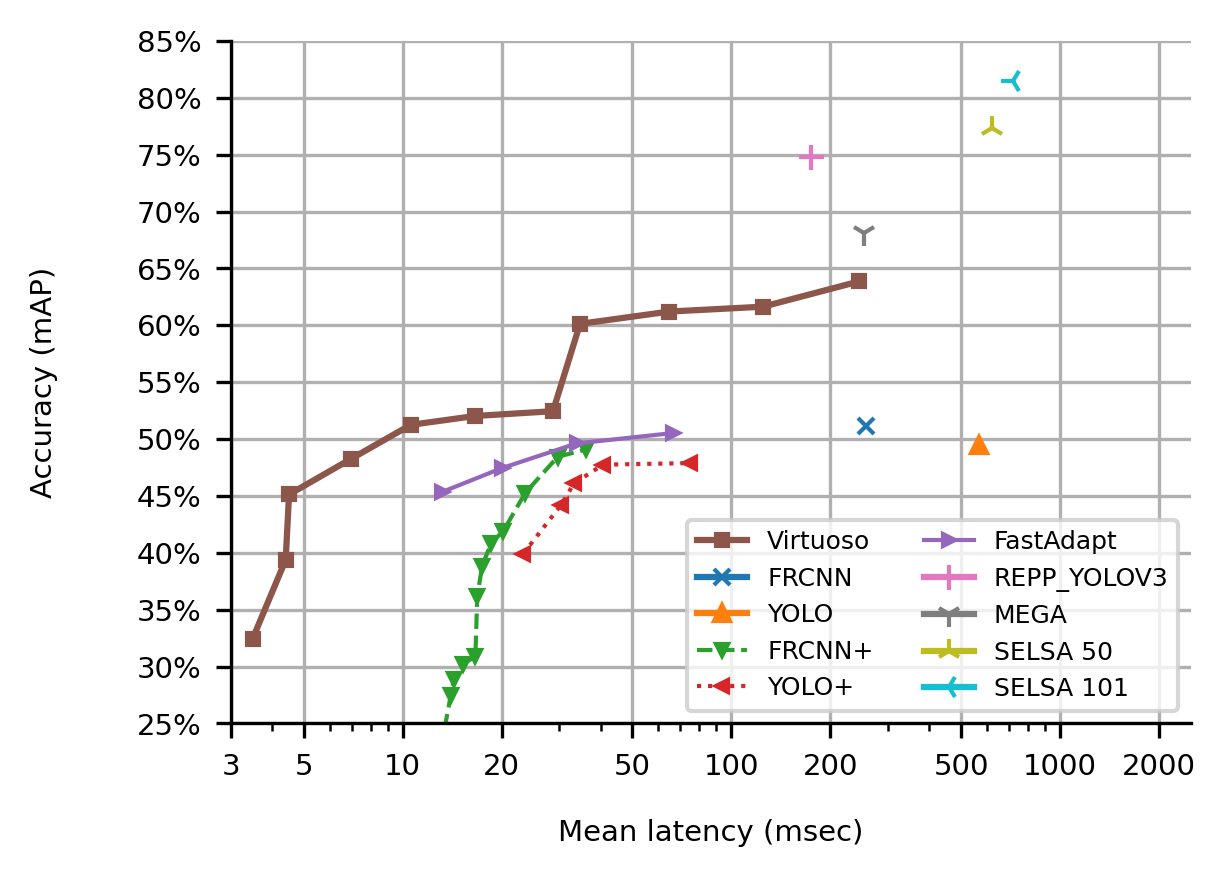}
    \caption{Comparison of \name with other baseline protocols including our in-house enhancements. \name is able to achieve a much wider range of adaptation with superior accuracy or latency tradeoff compared to all other protocols. Note that the x-axis is in log scale.}
    \label{fig:main_results}
    % \ran{Polish: change our name to \name.}
    % Jay 11/2 : Addressed. Also changed the legend column to 2.
\end{figure}

%%%%%%%%%%%%%%%%%%%%%%%%%%%%%%%
%%%%%%%   Figure 6, 7   %%%%%%%
%%%%%%%%%%%%%%%%%%%%%%%%%%%%%%%
\begin{figure}[t]
    \centering
    \begin{minipage}[c]{0.47\columnwidth}
        \includegraphics[width=1\textwidth]{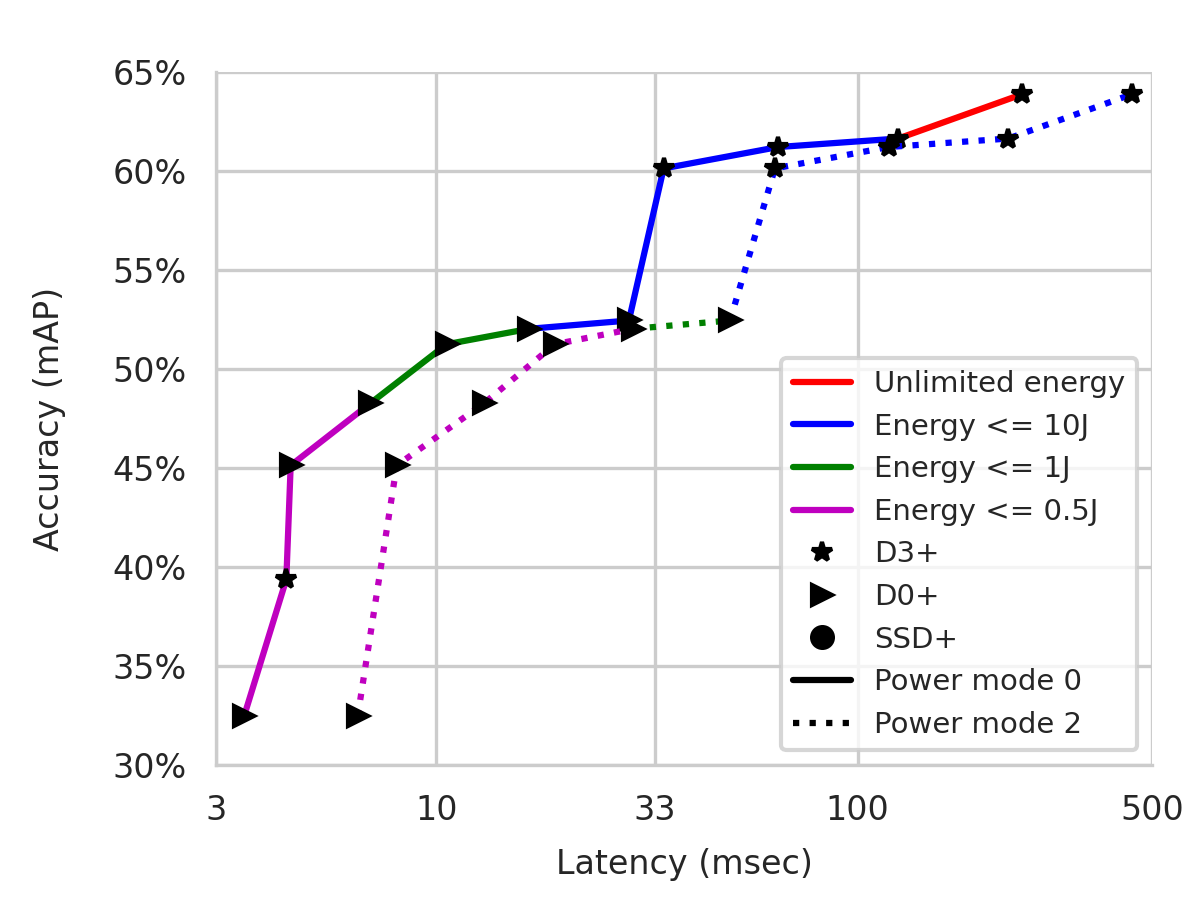}
        \vspace{-2em}
        \caption{Performance of \name's ability to adapt to different energy requirements under 2 different power modes, on NVIDIA AGX Xavier. }\label{fig:eng_req}
        % \ran{Polish: (1) legends change to "Energy <= 10 J", (2) Smaller fontsize by 1 for the legend, (3) legend name: Power mode, capitalize the first character.}
        % Jay 11/2 : Addressed
    \end{minipage}
    \hfill
    \begin{minipage}[c]{0.47\columnwidth}
        \includegraphics[width=1\textwidth]{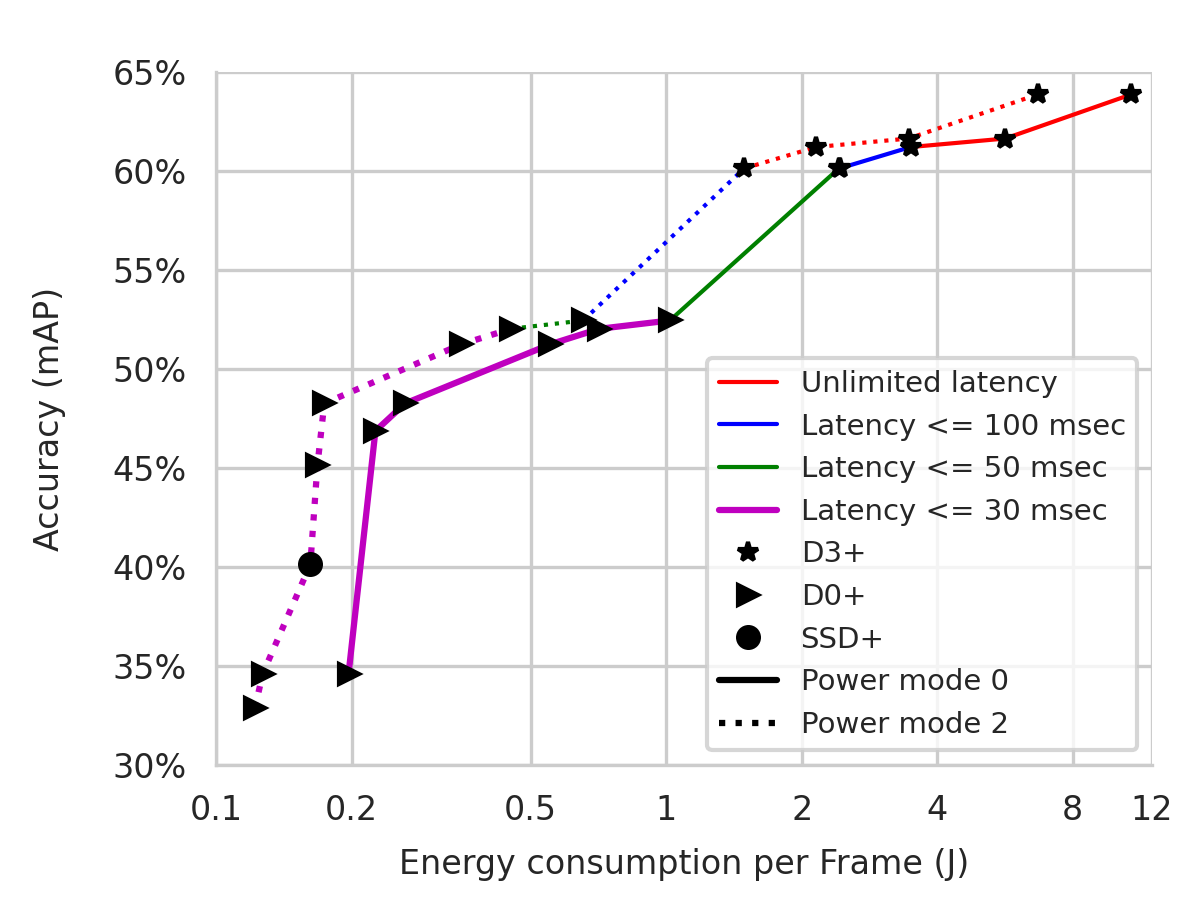}
        \vspace{-2em}
        \caption{Performance of \name's ability to adapt to different latency requirements under 2 different power modes, on NVIDIA AGX Xavier.}\label{fig:lat_req}
        % \ran{Polish: (1) legends change to "Latency = 100 msec", (2) Smaller fontsize by 1 for the legend, (3) legend name: Power mode, capitalize the first character, (4) change the fourth latency requirement to 33.3 msec which is more meaningful.}
        % Jay 11/2 : Addressed
        % \pengcheng{put other protocols as points here}
        % Jay 11/2 : Seems to make the plot too much crowded, and the for fig 6 and 7, we mainly focus for \name's ability to adapt on different power modes and requirements, so commenting for now.
    \end{minipage}
\end{figure}
%YL112321: Switch the captions for Fig 6 and 7?

% Jay: 10/27 Add content for the new figure 5 (main result figure)
We first present the overall evaluation results of \name and other efficient and adaptive video object detection baselines in Fig.~\ref{fig:main_results}.
% Jay : TODO : To be updated after updating figure 5. (Adascale + other baselines.)
% First observation: higher accuracy at the same latency requirement.
\textit{First}, \name achieves higher accuracy at any latency range between 15 msec to 200 msec. 
% \name
% lats = [3.5, 4.4, 4.5, 7.0, 10.6, 16.6, 28.7, 34.7, 64.7, 124.9, 245.3]
% accs = [32.5, 39.4, 45.1, 48.3, 51.3, 52.0, 52.5, 60.1, 61.2, 61.6, 63.9]
% FastAdapt  (50.5%, 66.3 msec), (45.4%, 13.2 msec)
% FRCNN+ (49.1%, 36.2 msec)
% YOLO+ (47.9%, 75.0 msec), (39.9%, 23.3 msec)
% FRCNN (51.1%, 257.0 msec)
% YOLO (49.5%, 566.0 msec)
%YL112321: The claim is fine, yet we have to admit that our method underperforms REPP for at least one latency requirement. Also, do we still plan to add adascale here?
Particularly, we are 10.7\% and 5.9\% more accurate than FastAdapt at 70 msec and 15 msec latency requirement, 11.0\% more accurate than FRCNN+ at 40 msec latency, 13.3\% and 12.1\% more accurate than YOLO+ at 80 msec and 25 msec latency, 12.8\% more accurate and 11.7 msec faster than FRCNN with maximum performance, and 14.4\% more accurate and 320.7 msec faster than YOLO, again, with maximum performance.
% \name 241.8 msec, FastAdapt 53.1 ,FRCNN+ 23.2, YOLO+ 51.7
\textit{Second}, \name achieves much wider latency adaptation range from 3.5 to 245.3 msec in which \name leads the accuracy frontier. The adaptation range is 4.5 times larger than FastAdapt, 10 times larger than FRCNN+, and 4.7 times larger than YOLO+.
To conclude, \name is able to achieve both superior accuracy and latency while covering a wider range of adaptation, outperforming all other efficient and adaptive baselines.

% 30 FPS branch.
% [['EFFDETD3', -1, -1, 8, 'medianflow', 2, 15, 'agx0'], [0.7518, 0.6013, 36.580813475, 2.419, 0]]
% [['EFFDETD0', -1, -1, 2, 'medianflow', 4, 15, 'agx0'], [0.7319, 0.5245, 28.66007264, 1.019, 0]]
We further examine the performance of \name given combined latency and efficiency requirements and under different power modes. We show in Fig.~\ref{fig:eng_req} a more detailed performance of \name given several energy requirements. We can see that given the most stringent energy requirement---0.5 J per frame, \name is able to achieve between 3.5 and 7.0 msec latency, with 32.5\% to 48.3\% accuracy accordingly (purple and solid curve). Then, we gradually relax the energy requirement to 1 J (green curve), 10 J (blue curve), and unlimited (red curve) and find that \name achieves higher accuracy (51.2\% to 63.9\%), at the expense of higher latency (10.6 msec to 245.3 msec). At a more real-world applicable frame rate of 30 FPS (33.3 msec per frame latency requirement), \name is able to run near 30 FPS (at 36.6 msec per frame) with an accuracy of 60.1\%.
% , or 28.7 msec per frame with an accuracy of 52.5\%. 
%
% On the other hand, the low latency and low accuracy performance given each energy requirement is the same.
% \somali{Can you reword this line above, unclear.}
% Jay 11/2 : Commented.

Therefore, we explicitly show the improvement in accuracy performance at a higher energy requirement with different colors. 
%
% 7.0 vs 29.3 (4.2 times)/ 16.6 vs. 49.8 (3 times) / 124.9 vs. 447.1 (3.6 times)
Furthermore, as we switch the power mode to 2 (from power mode 0) (dashed line), \name is more power efficient and achieves up to 51.2\%, 52.0\%, and 63.9\% accuracy with more relaxed energy requirements of 0.5 J, 1J, and 10 J. The accuracy is higher than that in power mode 0 given the same energy requirement, at the expense of 3 to 4.2 times higher latency.
%SC111921: I do not see this. I see accuracy for power mode 0 is higher than that for mode 2. 
% Jay 11/19 : While the overall plot of pm0 is above pm2, the plot is basically just a shift in the x direction, and when you compare with lines with the same color in fig6,(= branches with same energy requirements) blue for example reaches the highest accuracy in pm2, while for pm0, due to the energy requirement, the blue line reaches the point below the maximum accuracy.
%
One thing to note that is SSD+ is always inferior to EfficientDet+ in this experiment and thus has no data point in the figure (recall we are plotting the Pareto optimal curve). This is somewhat expected as EfficientDet is a more recent work with further optimized accuracy performance. Further, the lower latency region is dominated by EfficientDet D0+ and the higher accuracy region is dominated by EfficientDet D3+.
To conclude, the multi-requirement design of \name gives the flexibility to the user for picking different energy or latency requirements to match the use case and \name can always maximize its accuracy subject to such efficiency requirement.
%SC103121: What's meant by "efficiency requirement"? "Latency requirement"?
% Jay 11/1 : Yes, both energy and latency requirements

% Ran (10/27/2021): old description on Fig.~\ref{fig:eng_req}.
% We find that the execution branches, which are selected for a certain energy requirement, are always a subset for the branches, which are selected for a higher energy requirement. To highlight the branch selection difference made with different energy requirements, we only color the newly selected branches that are not included in a lower energy requirement. Our system can extend down to an extreme low latency of 3.5 msec with an accuracy of 32.5\% with power mode 0, and latency of 6.5 msec with accuracy of 32.5\% with power mode 2. Note that the accuracy performance is deterministic for a selected configuration, and the same execution branches were selected for both power mode 0 and power mode 2. For the higher accuracy performance range, \name can reach up to an accuracy of 63.9\% with 245.3 msec latency for power mode 0 and latency of 447.2 msec for power mode 2. 
% There is no single execution branch that can outperform all, and for example, under the same energy requirement of 10 J per frame, in Fig.~\ref{fig:eng_req} with power mode 0, \name is able to achieve up to 61.6\% accuracy with the latency at 124.9 msec. In contrast, \name is able to achieve the maximum performance of 63.9\% at the cost of higher latency of 447.2 msec with power mode 2. 

% Results for figure 7
We then evaluate \name with the latency requirement as a prioritized one and check the accuracy vs. energy tradeoff. Fig.~\ref{fig:lat_req} shows the adaptation performance of our overall framework for the accuracy vs. energy tradeoff under different latency requirements. While the results for accuracy vs. energy tradeoff shows a similar trend to results of accuracy vs. latency, one note is that SSD+ shows up in the Pareto performance curve in the power mode 2. This indicates that SSD+ is the preferred choice for some low energy budget cases due to its high energy efficiency and no object detector can dominate all users' requirements. For all models, as the latency requirement is made stricter, the overall energy consumption of selected branches also decreases, showing a generally applicable relationship between latency and energy consumption.

%%%%%%%%%%%%%%%%%%%%%%%%%%%%
%%%%%%%   Figure 8   %%%%%%%
%%%%%%%%%%%%%%%%%%%%%%%%%%%%
\begin{figure}[t]
    \centering
    \includegraphics[width=0.45\textwidth]{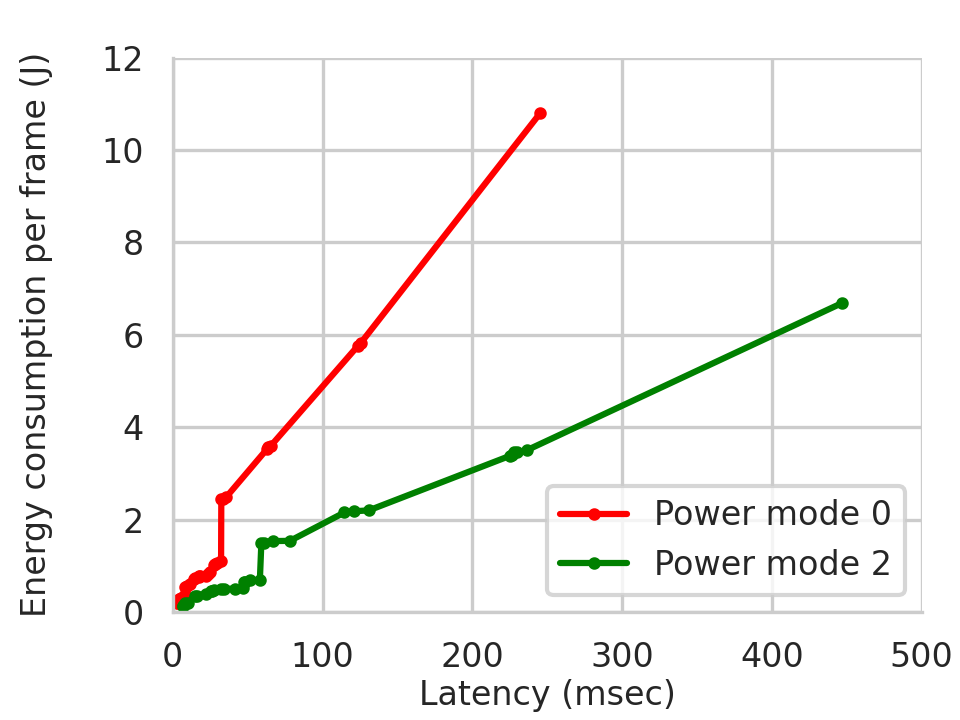}
    \caption{Energy vs. Latency performance for different power modes, on NVIDIA AGX Xavier. The execution branches that are selected as datapoints in each power mode lines are identical, and the difference in the power mode results in a proportional change between latency and energy consumption.}\label{fig:eng_vs_lat}
\end{figure} 

% SC101821: I don't understand: "the accuracy is identical across the same execution branch for each power modes".
% Jay 10/19 : for fig 7, the red line and green line is kind of 'symmetric' along the y=x line. They consist of identical execution branches - e.g, if power mode 0 has branch a,b,c power mode 2 has the same branch a,b,c with the performance symmetric to the y=x line(approximately). The power modes have a very constant impact on the energy / latency, so it is possible to observe a similar trend plot for the two power modes.
% Ran (10/27/2021): revised and see if it is clearer.

% \ran{Fill in XXX.}
% Jay 11/1 : Addressed.
To better understand the energy and latency performance of \name under different power modes, we show in Fig.~\ref{fig:eng_vs_lat} a more in-depth results where we select 31 execution branches from the multi-branch execution kernel of \name and show their energy consumption and latency. While the energy vs. latency tends to have a linear relationship, it is also observable that the power mode is an important factor that impacts both energy and latency. Roughly, power mode 2 is 40\% lower in energy consumption, with a drawback of 1.8 times higher latency compared to power mode 0.

%%%%%%%%%%%%%%%%%%%%%%%%%%%%%%%%
%%%%%%%   Figure 9, 10   %%%%%%%
%%%%%%%%%%%%%%%%%%%%%%%%%%%%%%%%
% Third figure with pareto per each model, accuracy vs. latency / accuracy vs. energy
\begin{figure}[t]
    \centering
    \begin{minipage}[c]{0.48\columnwidth}
        \includegraphics[width=1\textwidth]{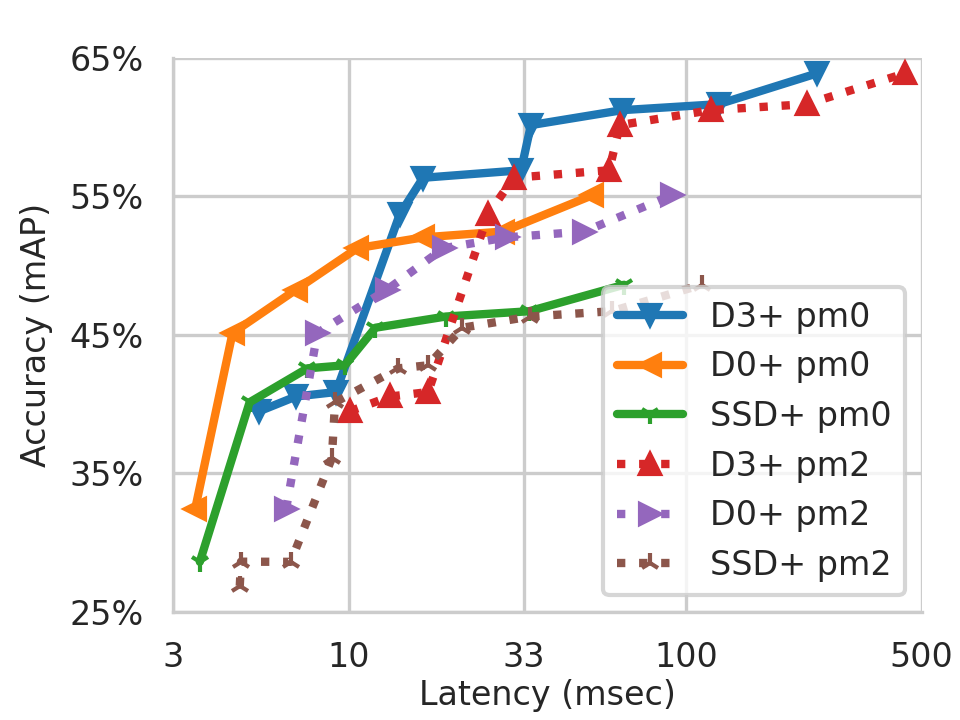}
        % \ran{Polish: "ssd+" --> ``SSD+''.}
        % Jay 11/2 : Addressed
        \caption{Accuracy of \name's each object detector backbone given latency requirements, on NVIDIA AGX Xavier.}\label{fig:separate_pareto_lat}
    \end{minipage}
    \hfill
    \begin{minipage}[c]{0.48\columnwidth}
        \includegraphics[width=1\textwidth]{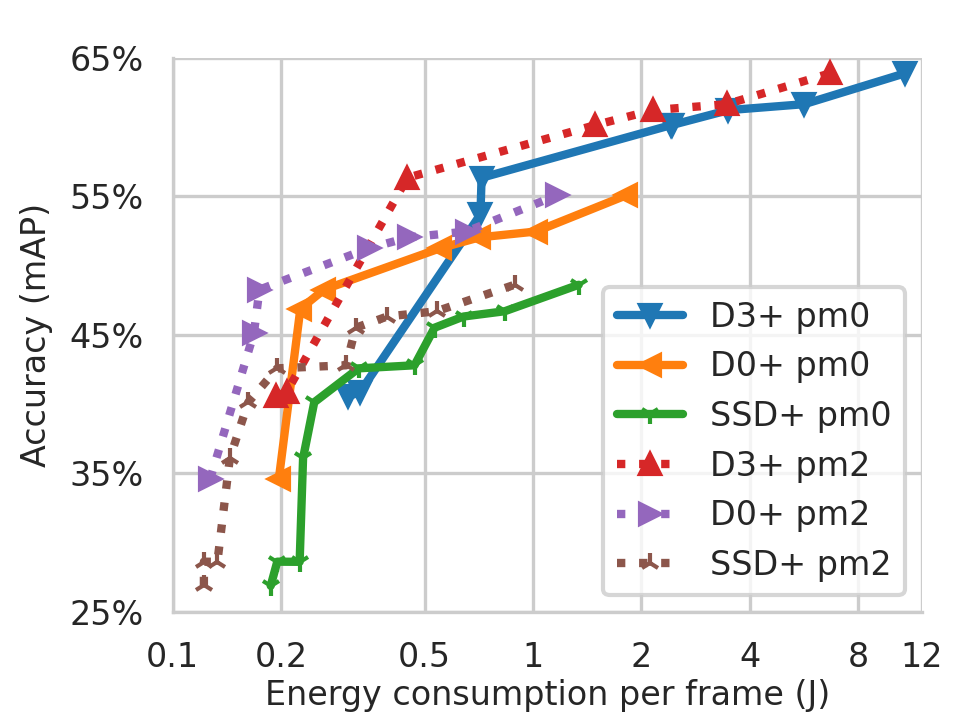}
        % \ran{Polish: "ssd+" --> ``SSD+''.}
        % Jay 11/2 : Addressed
        \caption{Accuracy of \name's each object detector backbone given energy requirements, on NVIDIA AGX Xavier.}\label{fig:separate_pareto_eng}
    \end{minipage}
\end{figure}

% Jay : reworded for fig 9, 10
% purpose : More detailed figure for showing performance of d0,d3,ssd
We dig deeper by examining the performance of each object detector backbone. Fig.~\ref{fig:separate_pareto_lat} and ~\ref{fig:separate_pareto_eng} show the performance of D3+, D0+ and SSD+ with accuracy vs. latency and accuracy vs. energy respectively. 
% Main observation : High impact of efficiency knobs.
All models show a similar trend of decreasing accuracy with better energy and latency performance, due to the object tracker related efficiency knobs being more dominant to enhance the efficiency.
For EfficientDet D3, the original model performs at 63.9\% accuracy with 245.3 msec latency and 10.8 J energy consumption per frame in power mode 0. This is reduced down to 5.4 msec and 0.3 J per frame which is 45 times faster, and 97.2\% more energy efficient at the cost of lower accuracy of 39.5\%. Similarly, with power mode 0, EfficientDet D0 is 8 times faster and 80.6\% more energy efficient with the accuracy dropping from 52.5\% to 32.5\%. For SSD, it is 9.5 times faster and 77.6\% more energy efficient. 
% Side observation/ : High impact of si on extreme low end.
Among our efficiency knobs, the detector interval has the highest impact on both the energy consumption and latency, since the object tracker is light-weight in both energy and computation cost, and also does not utilize the GPU.
For the extreme low end for each latency or energy consumption requirements, it is observed that the performance of SSD+ and EfficientDet D0+ are almost identical (D0+ has a slight advantage). This results in similar latency or energy consumption among execution branches at low rate of invocation of object detector.
For cases with the detector running more often, the object detector variant also plays a big role in the energy consumption and the latency. The deepest and most complex model D3+ has the highest energy consumption and latency, and the most light-weight model SSD+ has the lowest energy consumption and latency.
% Conclusion/takeaway : again emphasis the benefit of efficiency knobs that boost the latency/energy efficiency. Also the impact of power mode.
To conclude, our efficiency knobs not only benefit energy efficiency, but also improve latency performance. We also show that the power modes on embedded devices can further optimize the latency and energy performance.
%SC101821: Check above to get an example of what to emphasize as a takeaway lesson.

\subsection{Evaluation on the Accuracy and Latency across Devices}\label{sec:accuracy_latency_benchmark}

We further evaluate the performance of \name's each object detector backbone on more embedded devices --- NVIDIA AGX Xavier, Xavier NX, and TX2, and evaluate the effect of resource contention.
%SC102021: I don't think evaluating with contention is standard practice.
% Jay 10/25: The above sentence was intended to say that evaluating on the ILSVRC val set is a standard practice.
% Ran (10/30/2021): re-worded. No need to mention the dataset since we introduced in Section 5.2. Instead of saying sth is a standard practice, just say the PURPOSE of the experiment.

% Add description for scheduler under contention.
% \ran{Suggest to relocate/comment. This paragraph is only needed if weird results are observed to explain. The scheduler should have been aware of the contention.}
% Jay 11/1 : I see. Commented.
% For EfficientDet D0+, D3+, and SSD+, which are the object detector backbones in \name, the execution branches are determined by the scheduler under no contention. However, under contention, the latency performance is impacted and our prediction models for the scheduler does not work under such conditions. We manually re-use the execution branches selected by the scheduler under no contention, for experimenting on 50\% contention.

%%%%%%%%%%%%%%%%%%%%%%%%%%%%%%%%%%%%%
%%%%%%%   Figure 11, 12, 13   %%%%%%%
%%%%%%%%%%%%%%%%%%%%%%%%%%%%%%%%%%%%%
%SC102021: A general rule of thumb is a figure should not come so much before it is referred. I moved the figures later. 
% Jay 10/27 : change the order of the following three figures to match the order of ref.
\begin{figure*}[t]
    \centering
    \begin{minipage}[c]{0.45\textwidth}
        \includegraphics[width=1\columnwidth]{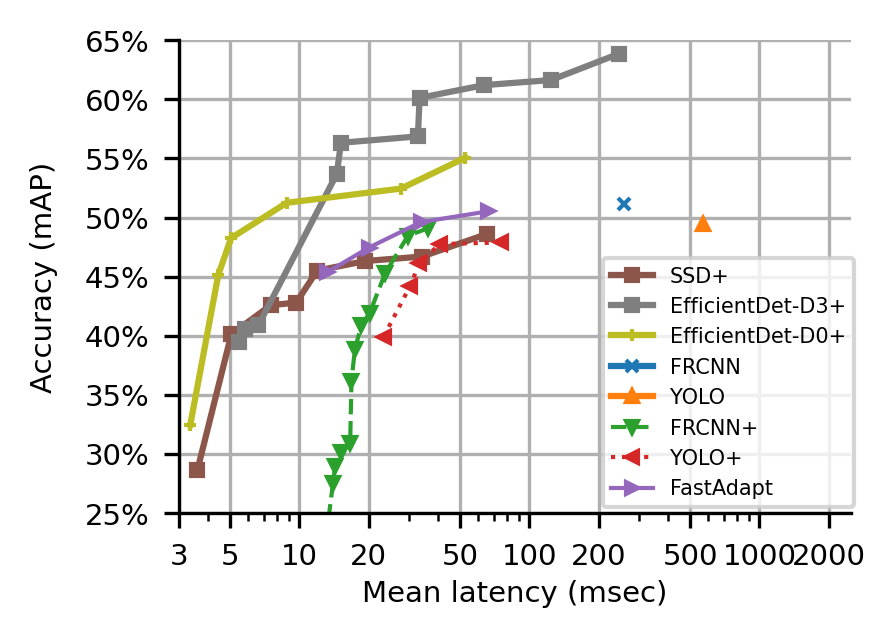}
        \centerline{\small (a) No contention}
    \end{minipage}
    % \hfill
    \begin{minipage}[c]{0.45\textwidth}
        \includegraphics[width=1\columnwidth]{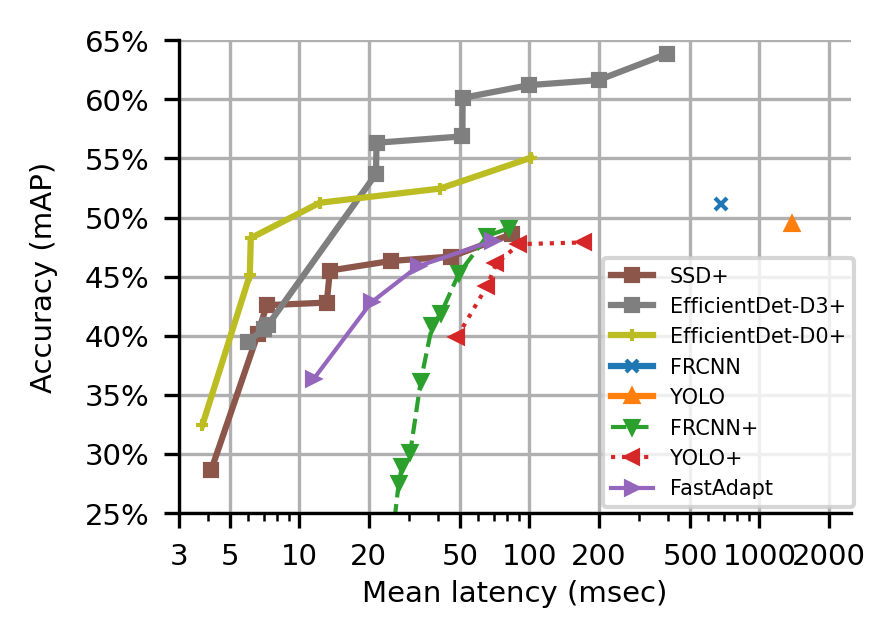}
        \centerline{\small (b) 50\% GPU contention}
    \end{minipage}
    \caption{Evaluating the object detector baselines on the NVIDIA Jetson AGX Xavier.}\label{fig:accuracy_latency_AGX}
  %  \ran{Polish: (1) Use code to save the figures into PNG files instead of doing screenshot. (2) Remove the large white space on the left and bottom.}
   % \jay{polished.} 
\end{figure*}

\begin{figure*}[t]
    \centering
    \begin{minipage}[c]{0.45\textwidth}
        \includegraphics[width=1\columnwidth]{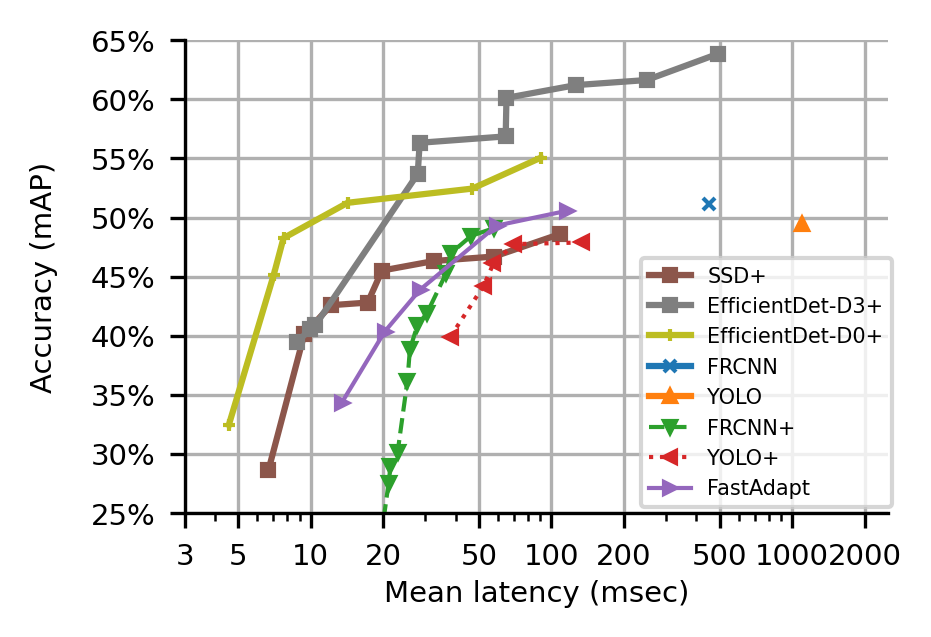}
        \centerline{\small (a) No contention}
    \end{minipage}
    % \hfill
    \begin{minipage}[c]{0.45\textwidth}
        \includegraphics[width=1\columnwidth]{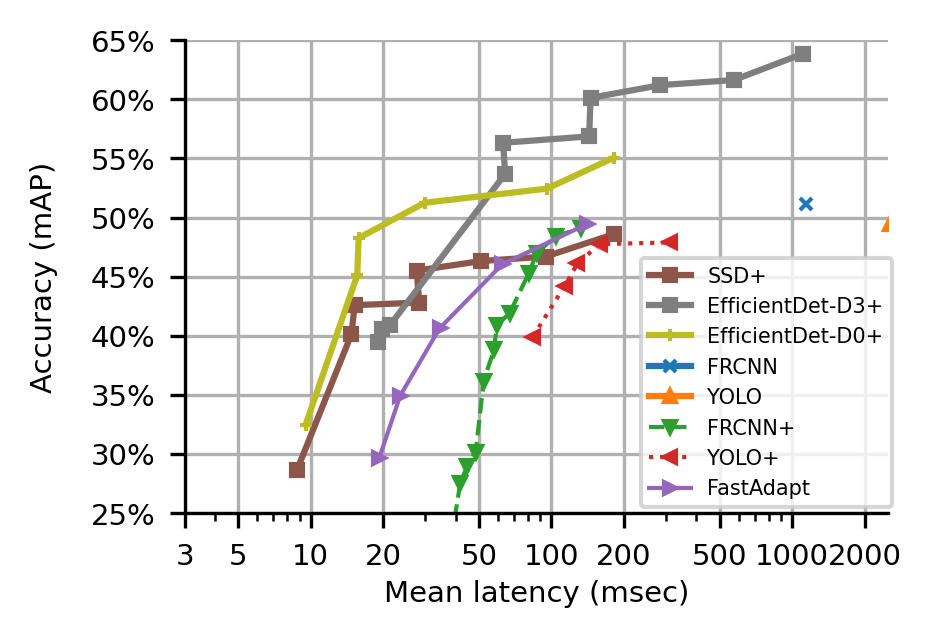}
        \centerline{\small (b) 50\% GPU contention}
    \end{minipage}
    \caption{Evaluating the object detector baselines on the NVIDIA Jetson Xavier NX.}\label{fig:accuracy_latency_NX}
\end{figure*}

\begin{figure*}[t]
    \centering
    \begin{minipage}[c]{0.45\textwidth}
        \includegraphics[width=1\columnwidth]{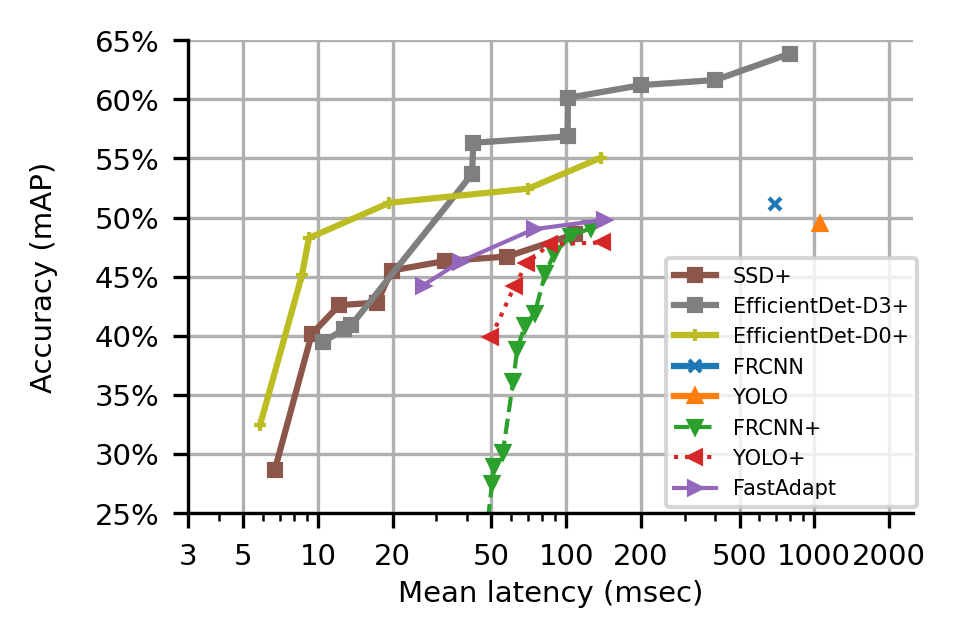}
        \centerline{\small (a) No contention}
    \end{minipage}
    % \hfill
    \begin{minipage}[c]{0.45\textwidth}
        \includegraphics[width=1\columnwidth]{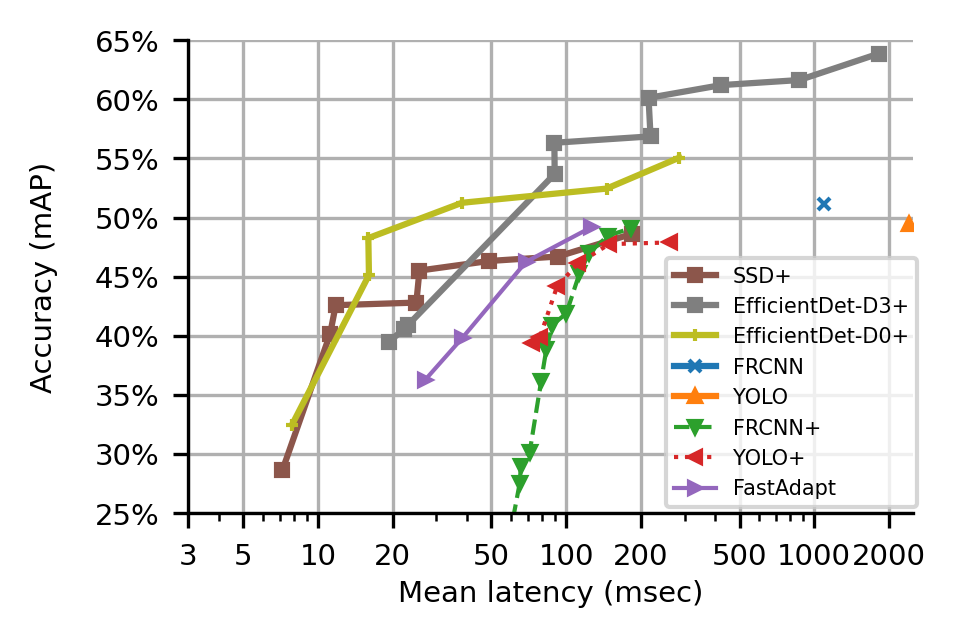}
        \centerline{\small (b) 50\% GPU contention}
    \end{minipage}
    \caption{Evaluating the object detector baselines on the NVIDIA Jetson TX2.}\label{fig:accuracy_latency_TX2}
\end{figure*}

\subsubsection{Adaptive Video Object Detection Models} \hfill\\

% purpose
% Main observation. Faster latency == better accuracy performance given the same latency requirement. Also focused mostly on d0+/d3+ and ssd+ being able to match strict latency requirements both with and without contention.
%\ran{real-time latency requirement is 33.3 msec, not 30. Why branches with 15.1 and 13.6 msec latency are chosen for meeting 33.3 msec requirment. There are no other branches?}
%\jay{yes, there are no other 'in-between' branches on the pareto for the given requirements.}
% Ran: got it. Thanks!

Fig.~\ref{fig:accuracy_latency_AGX} reports the accuracy and latency of all adaptive video object detection models with varying contention levels on the NVIDIA Xavier AGX board. 
% \ran{What is additional value of the AGX figure over the main Fig. 5 or Fig. 9? Tell the delta.}
% Jay 11/2 : Addressed by adding the sentence below.
Here, we also provide a more in-depth result of \name, by separately evaluating the multi-branch object detection kernels of \name - EfficientDet D0+, D3+ and SSD+.
% main observation: good performance of d0 and d3. Also d0, d3, and ssd which are all \name's variants have better performance over the baselines.
% \ran{Does SSD+ have the lead?}
% Jay 11/2 : No. Shows slight lead only in 50% contention on TX2 compared to d0+, but no.

% \ran{``our object detector backbones for \name'' means D0, D3. This is confusing. Consider \name variants.}
% Jay 11/2 : Modified to 'multi-branch object detection kernels of \name'.

\textit{First}, in subfigure (a), we observe that EfficientDet D0+, D3+, and SSD+, which are \name variants, have the lead in the accuracy-latency frontier over FastAdapt, FRCNN+, and YOLO+, over a wide range of latency performance, varying from 5.8 msec to 245.3 msec.
Specifically, EfficientDet D0+ has an accuracy up to 55.1\% mAP, running at 52.8 msec per frame (roughly 18.9 FPS) and has an latency down to 3.4 msec per frame (roughly 294.1 FPS), running at an accuracy of 32.5\% mAP.
% Mention FastAdapt here slightly, only by comparing with D0+. D0+ already outperforms FastAdapt by itself.
The maximum performance of EfficientDet D0+ is 5.3\% higher in accuracy compared to FastAdapt with the maximum performance, while having lower latency of 13.5 msec at the same time.
EfficientDet D3+ covers the higher accuracy range with higher latency compared to D0+. It can achieve up to an accuracy of 63.9\% mAP, running at 245.3 msec per frame (roughly 4.0 FPS), and the latency can be reduced down to 5.5 msec (roughly 181.8 FPS) with an accuracy of 39.5\%, combined with \name's efficiency knobs.
For SSD+, the accuracy is at maximum of 48.6\% with a latency of 65.5 msec per frame. The lowest achieved latency is 3.6 msec with an accuracy of 28.6\%, which is 0.2 msec slower and 3.9\% mAP lower compared to D0+.

% Side observation 1: YOLO+ / FRCNN+ is able to show better performance compared to YOLO/FRCNN
% Limited the comparison between FRCNN+ and YOLO+ since it is not the focus. only compare between its original model.
We also show the latency/accuracy performance of FRCNN+ and YOLO+ versus FRCNN and YOLO, the latter without our optimizations. For FRCNN and YOLO, the latency is 257 and 566 msec per frame while the accuracy is 51.1\% and 49.5\%, respectively. Coupled with our efficiency knobs, 
%
% Jay 11/2 : Tried to avoid presenting on low accuracy region, instead focus on 33.3 msec.
FRCNN+ is able to achieve a real-time processing time of 29.7 msec (33.6 FPS) at an accuracy of 48.4\%, which is more than 8 times faster than FRCNN, while only being 2.7\% lower in accuracy. Also FRCNN+ is able to achieve up to 49.1\% accuracy with 36.2 msec latency at maximum performance.
YOLO+ is able to achieve a minimum latency of 23.3 msec at 39.9\% accuracy, which is more than 24 times faster than YOLO. YOLO+ is also able to achieve 47.9\% accuracy at 75.0 msec with maximum accuracy performance.

% purpose : Show effect of contention
In Fig.~\ref{fig:accuracy_latency_AGX} (b), we measure the performance of these protocols under 50\% GPU contention.
%
% Main Observation:
% \ran{Cannot understand this observation.}
% Jay 11/2 : Added a more high-level observation statement.
Under 50\% GPU contention, most of the latency performance of each execution branches are increased by roughly 2 times, compared to no contention.
The D0+ branch that runs with 19.3 msec latency per frame under no contention, runs at 38 msec per frame with the same accuracy of 51.3\%. 
The D3+ branch that runs with 13.5 msec latency per frame under no contention runs at 23.0 msec per frame with 41\% accuracy.
%
% \ran{Is this AGX results? Be smart, do not say too much about extreme low accuracy region.}
% Jay 11/2 : chopped off and reworded.
Similar trend of increased latency is also observed for SSD+, FRCNN+, and YOLO+. The increased latency under contention reduces the accuracy performance under real-time processing latency of 30 FPS. SSD+ runs at 24.9 msec latency per frame with an accuracy of 46.3\%, and FRCNN+ runs at 30.4 msec latency per frame with 30.2\% accuracy. YOLO+ was not able to meet the 30 FPS latency requirement, but however, was still 28 times faster than YOLO with the fastest configuration, which is 48.7 msec, compared to YOLO's 1385.6 msec.
% With our efficient methods, FRCNN+ and YOLO+ are able to come down to a latency of 47.9 msec and 49.9 msec respectively, % with a very low accuracy of 22.6\% and 39.9\% respectively.

% Observation for NX board.
Fig.~\ref{fig:accuracy_latency_NX} shows the evaulation results on Xavier NX. Subfigure (a), given a latency requirement of 50 msec (20 FPS), EfficientDet D0+ achieves 52.5\% accuracy with 46.9 msec latency per frame. 
% Jay 11/2 : Change TX2 to AGX and update values.
This is 1.2\% higher in accuracy compared to the performance on TX2 (Fig.~\ref{fig:accuracy_latency_TX2} where EfficientDet D0+ achieves 51.3\% accuracy with 38.0 msec latency per frame, under the same latency requirement of 50 msec. Similarly, EfficientDet D3+ achieves 56.3\% accuracy with 28.4 msec latency on Xavier NX, where it achieves the same accuracy of 56.3\% with 42.2 msec latency on TX2 under 50 msec latency requirement. 
% \ran{Use 33.3 ms, which is meaningfull.}
% Jay 11/2 : Addressed.
Dialing up the latency requirement (more stringent and realistic for video) to 33.3 msec, the accuracy performance on TX2 drops to 40.9\% with 13.5 msec latency per frame.
%
% \ran{47\% accuracy should come with 1 digit after the decimal point.}
% Jay 11/2 : Addressed.
This accuracy increase is also observed with other baselines applied with our efficiency knob. Both FRCNN+ and YOLO+ cannot run under 40 msec latency requirement on TX2, but on Xavier NX, FRCNN+ can run with 47.0\% accuracy at a latency of 38.3 msec, and YOLO+ can run with 39.9\% accuracy at a latency of 38.1 msec.
We further evaluate the effect of GPU contention in Fig.~\ref{fig:accuracy_latency_NX} (b). We also observe increased latency of all baselines as the contention is given. However, EfficientDet D0+, D3+, and SSD+ can still keep the real-time 30 FPS (33.3 msec per frame) with the accuracy of 51.3\%, 40.9\%, and 45.5\% on Xavier NX when there is 50\% GPU contention. 
%\ran{Maybe wrong text? D0+ is more accurate than D3+?}
%\jay{These are correct. The D0+ is more accurate since the comparison is made on 30 FPS, where D0+ is above the D3+ pareto for NX. Also another reason would be that there are some gap between pareto branches, not having 'enough' middle points to handle specific requirement values.}
% Ran: got it. Thanks!
%
% Jay 11/2 Added another sentence to briefly address this.
Note that EfficientDet D0+ has a higher accuracy of 51.3\% compared to EfficientDet D3+'s 40.9\% under a stringent latency requirement. Such accuracy vs. latency tradeoff benefits \name with all multi-branch kernels combined, being able to switch between kernels to achieve an overall higher pareto curve.

% TX2
% Jay 11/2 : reworded the below sentence.
On the TX2 device, the performance trend of baselines are similar to performance on NX or AGX devices, only showing increased latency due to the device's weaker computation power. Comparing with results from Xavier AGX and NX boards, we observe that given a stronger computation power, our multi-branch object detection kernels of \name can achieve higher accuracy performance under the same latency requirement.

% main observation: good performance of d0 and d3. Also d0, d3, and ssd which are all \name's variants have better performance over the baselines.
Fig.~\ref{fig:accuracy_latency_TX2} shows that all baselines achieve roughly 2 times higher FPS on AGX Xavier compared to that on TX2. In addition, we were able to observe that a device with higher computation power gives larger benefits to execution branches with heavier object detector backbones, where EfficientDet D3+ is able to run at 53.7\% accuracy on AGX Xavier with 15.1 msec latency per frame, whereas it runs with 40.9\% accuracy at 13.6 msec latency on TX2 under a real-time latency requirement of 30 msec per frame.
%
% Note: FRCNN+ and YOLO+ low accuracy is due to limited implementation of our efficiency knobs.
%
% \ran{Weak argument: can we simply suggest not to use the low accuracy region? Why having to emphasize this part. Be smarter when claiming some credit of using the efficiency knobs and avoid the criticism of having low accuracy.}
% Jay 11/2 : Originally thought to defend the low accuracy region, but just commented. I agree that this makes a weak claim.
% Note that the FRCNN+ and YOLO+ come with very low accuracy at minimum latency, and this is mostly due to implementing a fixed number of 8 for the detector interval knob, rather than its full range.
%
% Jay 11/2 : relocated.
One thing to note is that, different multi-branch object detection kernels have slightly different capability of handling contention. While SSD+ is always inferior to EfficientDet D0+ under no contention (subfigure (a)), SSD+ comes close, or even surpasses D0+ in certain areas under 50\% contention, which also validates the benefit of having multiple kernels in \name.

% Jay 11/2 : Commented, since I think it has some overlapping with the overall conclusion after edit.
% % Conclusion for fig 11: even under contention, \name can adapt.
% \ran{Weird conclusion. Any design or contribution says \name can adapt? We should conclude with better accuracy. Conclusion should JUSTIFY our contribution and techniques.}
% To conclude, the three \name variants---D0+, D3+, and SSD+ still maintain the real-time processing latency of 33.3 msec (30 FPS) with reasonable accuracy under contention, upholding the claim of adapting to resource availability at runtime. 
 
%
% Jay 10/28 : Overall conclusion of fig 11~13
% \ran{Overall, the conclusion here is not interesting. Think about the experiment setting--multiple devices and multiple contention scenarios. What we want to evaluate and what has been achieved.}
% Jay 11/2 :Rewritten the conclusion
% 1. Good performance of multi-branch object detection kernels of \name
% 2. Different devices - leveraging of \name with different computation power.
% 3. Impact of contention
% Ran: Thanks.
To summarize, first, multi-branch object detection kernels of \name - EfficientDet D0+, D3+ and SSD+ - lead the accuracy-latency performance in all scenarios, especially under stringent latency requirements. 
% Jay 11/2 : chopped off below.
% all 3 groups of experiments suggest that as contention increased, the latency is affected and the accuracy drops accordingly to conform to the latency requirements. %
Compared to other adaptive video object detection baselines - FastAdapt, FRCNN+, YOLO+, EfficientDet D3+ can achieve up to 15.9\% better accuracy, 
Second, using different embedded devices shows that, given a stronger computation power, EfficinetDet D0+, D3+ and SSD+ can leverage higher accuracy under a specific latency requirement, due to the latency performance boost. In contrast, non-adaptive baselines only have a single datapoint of accuracy vs. latency and cannot leverage the accuracy with the changing latency performance.
Last, we were able to observe that our multi-branch object detection kernels of \name can still maintain real-time latency performance with reasonable accuracy. While GPU contention impacts the overall latency performance for all baselines regardless of the device being used, unlike non-adaptive baselines where the latency increased up to 2 times, EfficientDet D0+, D3+ and SSD+ were able to maintain real-time latency of 19.3 msec, 13.5 msec, and 23.0 msec on AGX device with 51.3\%, 41\%, 46.3\% accuracy respectively.

% \ran{Cannot understand the following sentence.}
% Jay 11/2 : Agreed. Seems a little irrelevant. Chopped off.
% EfficientDet D0+, D3+ and SSD+ is able to schedule the efficiency knobs with more options, especially with the adaptive detector interval that controls how often the object tracker is triggered. The object tracker uses much less GPU resources, compared to the object detector backbone, and suffers relatively lower accuracy loss for most cases especially in cases where the tracking of the objects is not lost. 
%
% \ran{Build more login connections between sentence. I cannot conclude the following sentence with the sentence above.}
% Jay 11/2 : Tried to address this by building 3 items for conclusion.
Our results show that \name, which leverages all three multi-branch kernels of EfficientDet D0+, D3+ and SSD+, has the ability to meet the latency requirement more flexibly under various scenarios while maintaining higher accuracy than other baselines.

%%%%%%%%%%%%%%%%%%%%%%%%%%%%%
%%%%%%%   Figure 14   %%%%%%%
%%%%%%%%%%%%%%%%%%%%%%%%%%%%%
\begin{figure*}[t]
    \centering
    \begin{minipage}[c]{0.32\textwidth}
        \includegraphics[width=1\columnwidth]{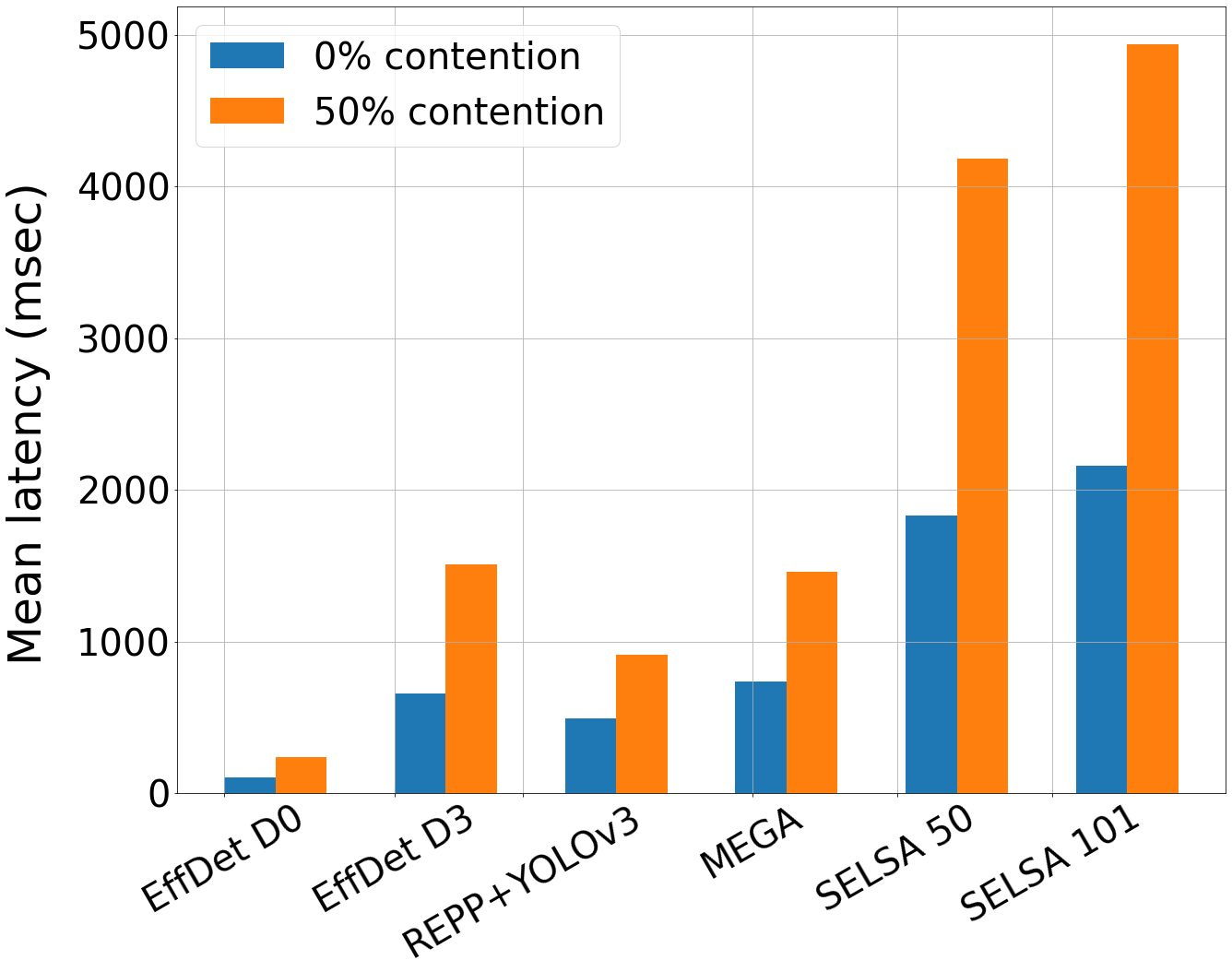}
        \centerline{\small (a) Jetson TX2}
    \end{minipage}
    \hfill
    \begin{minipage}[c]{0.32\textwidth}
        \includegraphics[width=1\columnwidth]{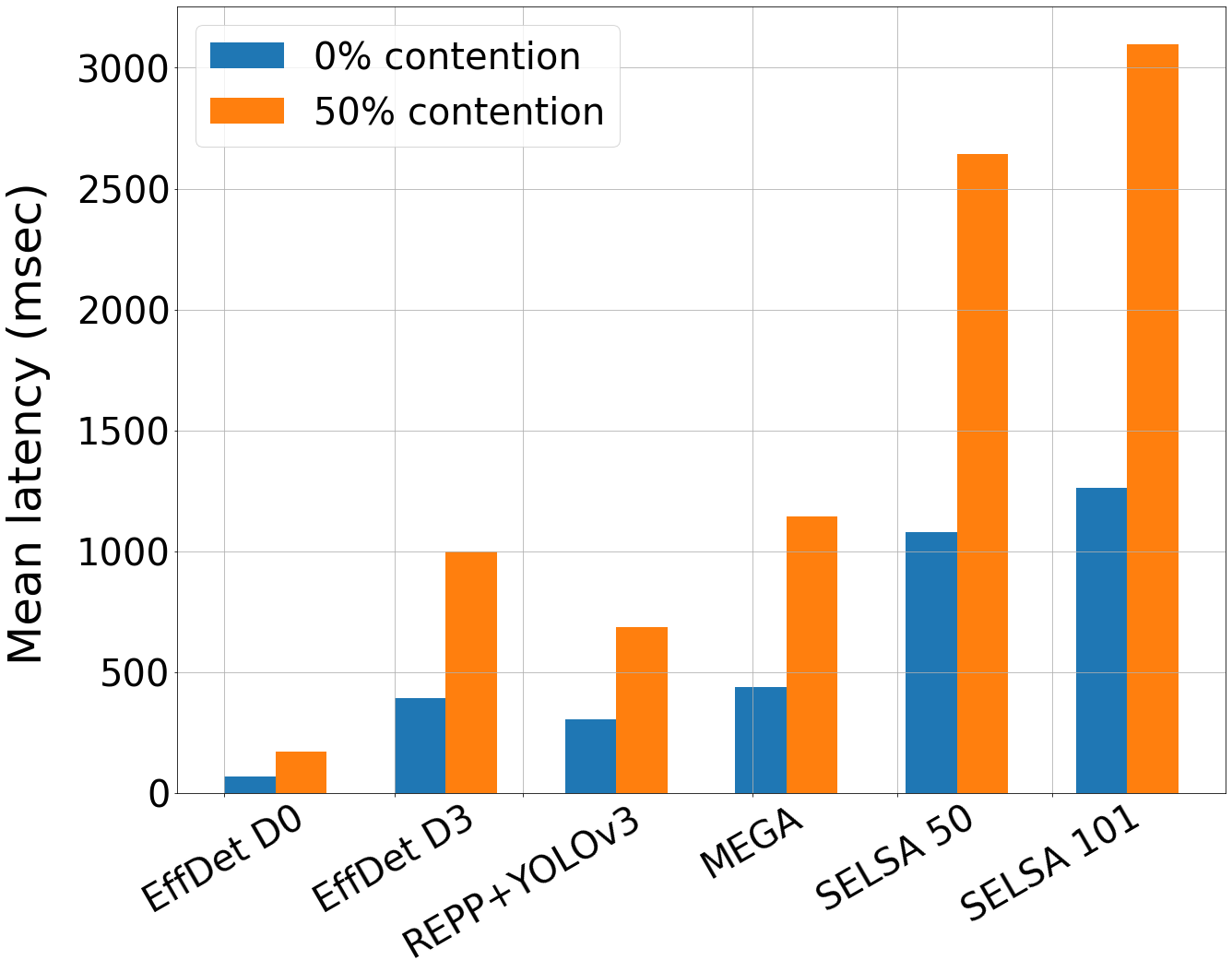}
        \centerline{\small (b) Jetson Xavier NX}
    \end{minipage}
    \hfill
    \begin{minipage}[c]{0.32\textwidth}
        \includegraphics[width=1\columnwidth]{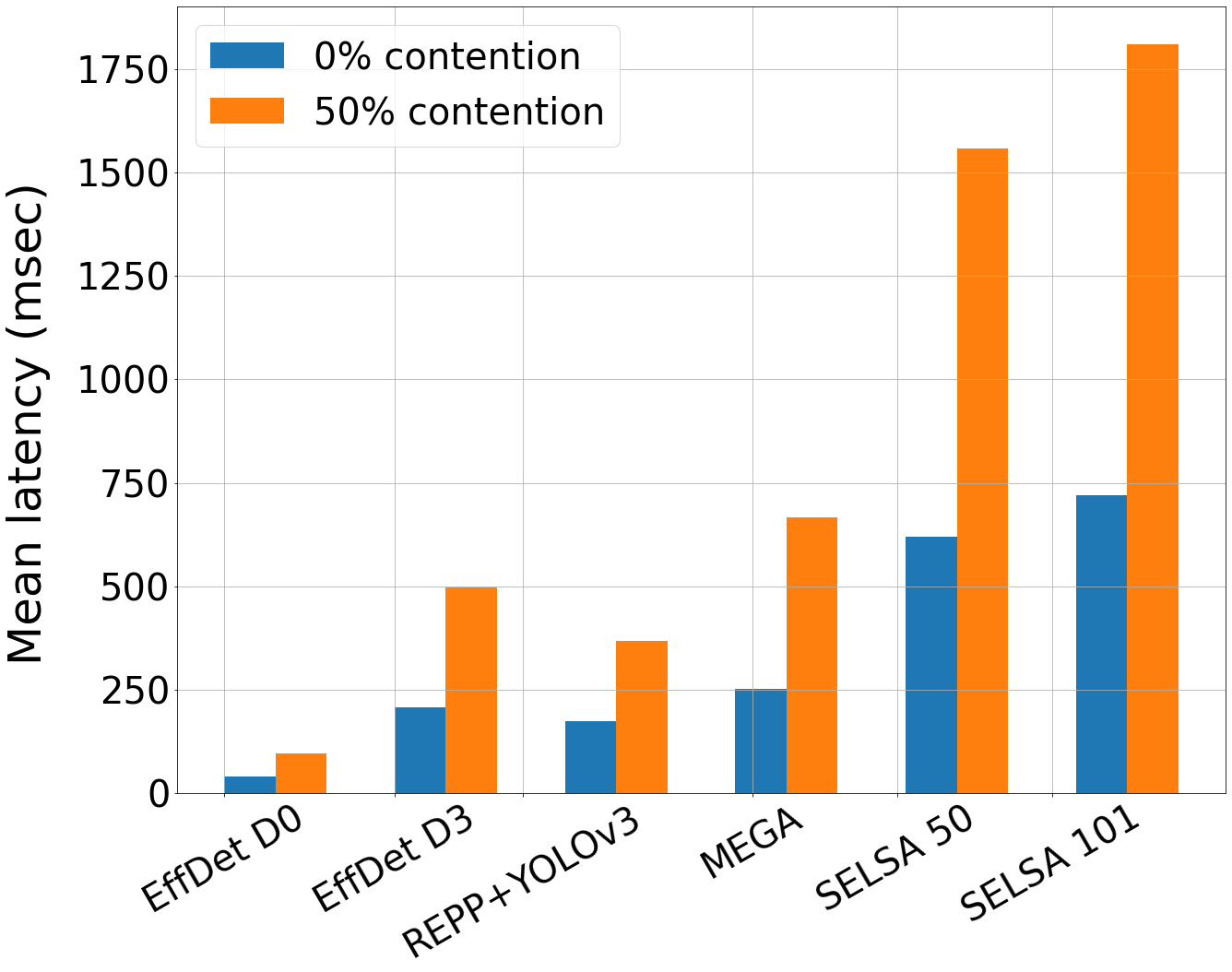}
        \centerline{\small (c) Jetson AGX Xavier}
    \end{minipage}
    \caption{Benchmarking SOTA video object detection models on NVIDIA Jetson Devices under different contention levels. The accuracy measurements are the same among all boards and are as follows. EfficientDet D0: 55.07\% mAP, EfficientDet D3: 63.87\% mAP, REPP+YOLOv3: 74.81\% mAP, MEGA base: 68.11\% mAP, SELSA 50: 77.31\% mAP, SELSA 101: 81.5\% mAP. Unlike the adaptive baselines, non-adaptive baselines provide only a single datapoint of accuracy and latency under the given environment.}\label{fig:sota}
    % \ran{Remove ``YOLOv3'' from figures.}
    % \jay{Updated caption, for the figure, to be removed.}
    % \somali{Similar to the comment earlier in this section, use a clearer term than MEGA base.}
    % Jay 11/2 : Addressed.
\end{figure*}

% Jay 10/28 : move all single datapoints here? Maybe add FRCNN, YOLO as well?
% \ran{Confusing caption. Are they non-adaptive?}
% Jay 11/2 : Yes they are non-adaptive, but since they are kinda grouped with FRCNN+ and YOLO+, decided not to add here.
\subsubsection{Accuracy-Optimized Video Object Detection Models} \hfill \\
% Both SOTA models (YOLOv3, REPP+YOLOv3, MEGA base, SELSA 50, 101) and efficient models (EfficientDet-D0, EfficientDet-D3) 

% purpose.
% \ran{Use the word ``evaluate'' more. Eliminate the word ``experiment''. Experiments are used to evaluate something. State your purpose--what aspect do you want to evaluate.}
% Jay 11/2 : Addressed.
% \ran{We should explicitly mention that we use D0 and D3 to compare with XXXs in the PURPOSE sentence.}
% Jay : 11/2 Addressed.
\noindent Baselines without adaptive features are evaluated for accuracy and latency on different embedded devices, and we use EfficientDet D0 and D3 as the base for comparison. 
%
% Main observation. : accuracy is higher for accuracy oriented models, but poor latency. Much worse under contention.
As we can see from the caption of Fig.~\ref{fig:sota}, without latency requirements, these baselines achieve higher mAP than most adaptive baselines, ranging from 55.1\% to 81.5\%. 
% \ran{What prior work?}
% Jay 11/2 : commented. I used prior work to point to our other results, which I also think is confusing. But seems it is just redundant.
For latency results, EfficientDet D0 has the lowest latency and accuracy among all models and boards.
% which is in line with our results. 
% 
SELSA 101 achieves the highest accuracy but with the highest latency on all boards. In contrast, REPP with YOLOv3 has a reasonable accuracy, at 74.81\%, while still maintaining a relatively low latency. However, when considering a user latency requirement, most baselines have a latency over 200 msec, or 5 FPS, even on the AGX Xavier board, which is one of the most embedded boards (subfigure (c)). Such latency values do not meet the need for real-time processing. We find from our evaluation that all baselines without any adaptive features suffer from poor latency, further exacerbated under GPU contention, showing up to 2X increased latency with contention. This is because of their larger network sizes, sophisticated design of using frame aggregation, as a post-processing step and inability to adapt to the runtime environment.
% \somali{as a post-processing step} 
%
% Note : Accuracy Inconsistency
The accuracy values obtained in our evaluation for non-adaptive baselines differ from those of the original authors, most significantly for REPP. This, we believe, is due to two reasons, as follows:
%\ran{Be clear. Is it IoU threshold during the NMS phase? Why choosing 0.6? Should we use the ``streaming'' setting to explain the difference? The second reason is hard to understand since it is too detailed.}
% Jay 11/2 : Addressed the IoU threshold. Replaced the second reason to point to streaming setting and limitation in frame aggregation.
% Ran: got it. Thanks!
First, we use an IoU threshold of 0.6 during the non-max suppression phase, consistently for all protocols which is mostly used as the default for our adaptive baselines (original authors use different thresholds for different protocols, \eg 0.5 for REPP). Second, we use a streaming setting of input frames, where the information to future frames that are required for frame aggregation of these non-adaptive baselines, is limited.
% and second, for the REPP evaluation, we use the code's ``demo'' version, rather than the ``full'' version, which has most parameter values hard-coded and thus not easy to use. 

% Overall Conclusion + takeaway for fig 11~14
% Jay 11/2 : Changed to conclusion for fig 14.
%\ran{What's the conclusion related to multiple devices?}
To conclude, we show in Fig.~\ref{fig:sota}, that the 50\% contention on the embedded device impacts the latency performance of each object detection kernel, and increases the latency by roughly two times. Also similar trend of better latency performance of using a device with a stronger computation power is observed as in Fig.~\ref{fig:accuracy_latency_AGX} to Fig.~\ref{fig:accuracy_latency_TX2}. 
% Jay 11/2 : Main claim for this short conclusion
However, unlike our adaptive kernels of EfficientDet D0+, D3+ and SSD+, non-adaptive baselines do not have any ability to meet the latency requirement, and shows a very high latency of over 300 msec, under 50\% GPU contention even on the Xavier AGX board.

% While the latency under contention increases, still our in-house implemented models are able to switch to a lower accuracy execution branch, and maintain real-time processing latency.

\subsection{Evaluation of Energy Consumption}\label{sec:energy-exp}

% \ran{Consider to move this section above the accuracy-latency evaluation.}
% Jay 11/2 : Considered, but since we only checked on the NX board, I think it is better to keep it here, even if we want to emphasize energy more in this paper.

Embedded devices need to be energy efficient as in most cases they may need to run on limited power. For a more in-depth profiling and comparison for energy consumption of different baselines with different power modes, we explicitly select the Jetson NX board as it is our middle-of-the-computational-power device and also comes with a moderate selection of power modes. 

% Jay 10/26 : Updated figure, add more caption.
\begin{figure}[th!]
    \centering
    \includegraphics[width=1\linewidth]{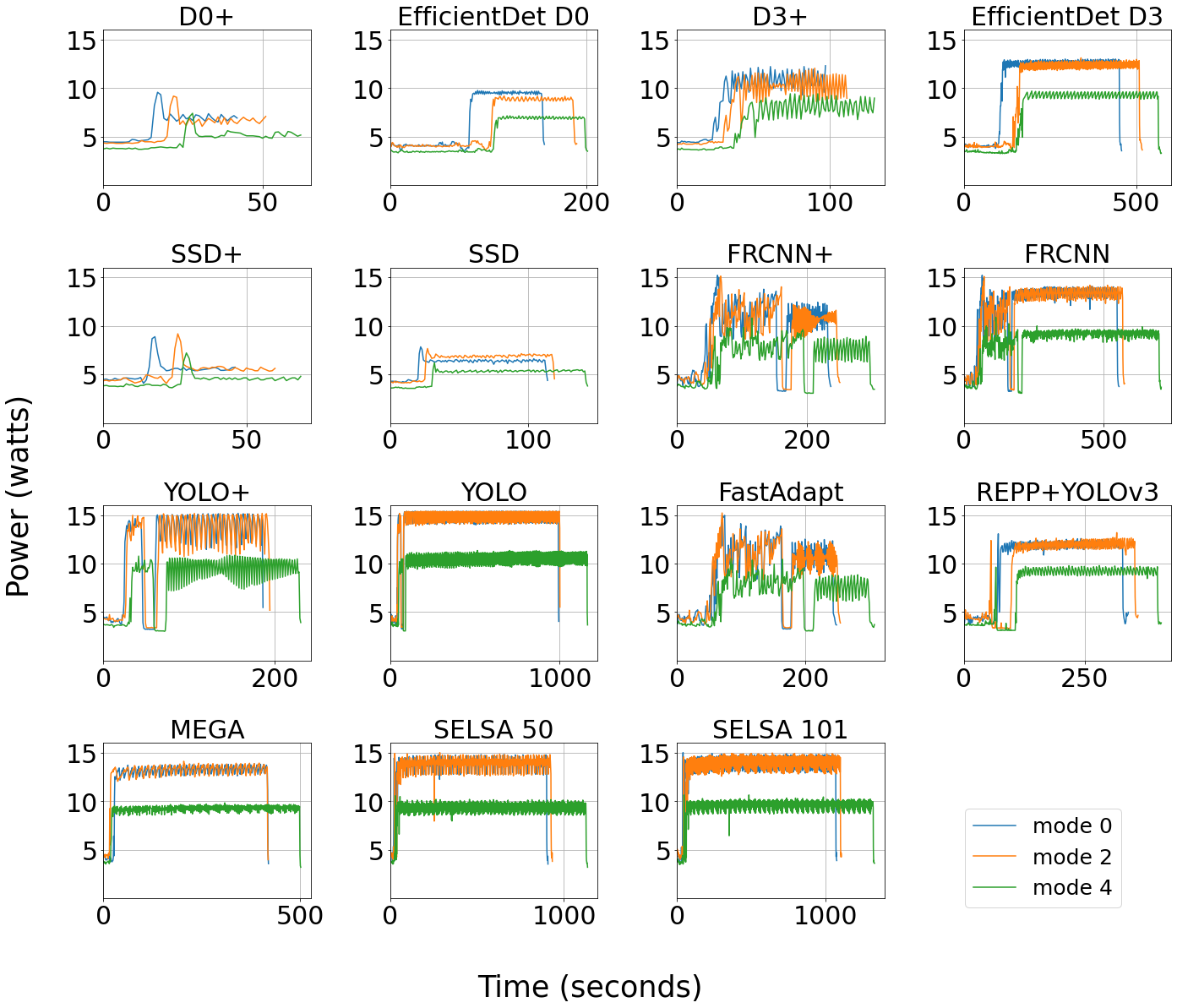}
    %\jay{Addressed, in a slightly different way. Also kept FastAdapt for now, need decision. TODO: Unify D0+, D0, D3+, D3 names.}
    \caption{Instantaneous power on Xavier NX. Overall, our in-house baselines with our efficiency knobs show a lower average power level compared to its original backbone without any efficiency knobs.
    Heavier and accuracy-oriented models show higher instantaneous power and longer inference times.}\label{fig:energy}
    % Jay 11/3 Relocated the previous caption to table 3, and updated.
    % \somali{Remember to change the name in the plots for our protocol to \name.}
    % Jay 11/2 Addressed.
\end{figure}

\begin{table}[th!]
\centering
\small
\scalebox{0.67}{
    \begin{tabularx}{1.5\columnwidth}{|>{\hsize=.40\hsize\centering\arraybackslash}X
    |>{\hsize=.40\hsize\centering\arraybackslash}X
    |>{\hsize=.40\hsize\centering\arraybackslash}X
    |>{\hsize=.40\hsize\centering\arraybackslash}X
    |>{\hsize=.45\hsize\centering\arraybackslash}X
    |>{\hsize=.45\hsize\centering\arraybackslash}X
    |>{\hsize=.45\hsize\centering\arraybackslash}X
    |>{\hsize=.45\hsize\centering\arraybackslash}X
    |>{\hsize=.45\hsize\centering\arraybackslash}X
    |>{\hsize=.45\hsize\centering\arraybackslash}X
    |>{\hsize=.45\hsize\centering\arraybackslash}X
    |>{\hsize=.50\hsize\centering\arraybackslash}X
    |>{\hsize=.53\hsize\centering\arraybackslash}X
    |>{\hsize=.45\hsize\centering\arraybackslash}X
    |>{\hsize=.45\hsize\centering\arraybackslash}X
    |>{\hsize=.45\hsize\centering\arraybackslash}X|}
    \hline
    \textbf{Power Modes} & \textit{D0+} & \textit{D0} & \textit{D3+} & \textit{D3} & \textit{SSD+} & \textit{SSD} & \textit{FRCNN+} & \textit{FRCNN} & \textit{YOLO+} & \textit{YOLO} & \textit{Fast Adapt} & \textit{REPP w\ YOLOv3} &  \textit{MEGA}  & \textit{SELSA 50} & \textit{SELSA 101}\\ 
    \hline
    \hline
    0 & 267 & 1050 & 882 & 4786 & 295 & 692 & 2205 & 6567 & 2196 & 14270 & 2244 & 3427 & 5269  & 12233  & 14565\\
    \hline
    2 & 305 & 1183 & 937 & 5050 & 346 & 757 & 2296 & 6745 & 2246 & 14407 & 2293 & 3490 & 5368 & 12519 & 14967 \\
    \hline
    4 & 297 & 1025 & 863 & 4323 & 347 & 716 & 2064 & 5906 & 1914 & 11836 & 2069 & 3122 & 4528 & 10493 & 12451 \\
    \hline
    Avg & 290 & 1086 & 894 & 4720 & 329 & 722 & 2189 & 6406 & 2119 & 13504 & 2202 & 3346  & 5055 & 11748 & 13995 \\
    \hline
    \end{tabularx}
}
%\jay{Updated table, following the same order with fig 15.}
% Jay 11/3 : Relocated caption from fig 15.
% \somali{ranging from XXX to XXX}
% Jay 11/3 : Internal purpose. FRCNN+ : 65%, D3+ : 81%
\caption{Energy consumption (J) on Jetson Xavier NX. Overall, our in-house baselines with our efficiency knobs show a huge efficiency boost in energy consumption ranging from 65\% to 81\% reduction, compared to the baselines without our efficiency knobs.}\label{tab:energy}
\end{table}

\noindent \textbf{Experiment setup}: With the selected power modes of Xavier NX from Section.~\ref{sec:embedded}, we first measure the power in the idle status. The results are 3.25, 3.38, and 3.08$W$ in mode 0, 2, and 4 respectively. Then, we run each video object detection baselines on a randomly selected video, which has 828 frames. The results are shown in Fig.~\ref{fig:energy} and Table~\ref{tab:energy}.
% From a total of five power modes shown in Table~\ref{tab:nx-power-modes}, we select power mode 0, 2, and 4, in our experiments to cover most usage scenarios, and observe the impact of different power budgets and processor configurations on energy consumption.

% Jay 11/2 : Commented
% \ran{What's the point of observing the power mode table? This is not our results. We should not observe. Refer to it and say which we use.}
% As can be seen from Table~\ref{tab:nx-power-modes}, power mode 0 provides the highest power budget (15W), with the highest frequency setting for each of GPU, CPU, and the deep learning accelerator; power mode 2 also provides a 15W power budget, with the CPU frequency setting lower than mode 0, and having 6 CPU cores enabled, compared to only 2 CPU cores in mode 0. Power mode 4 provides the lowest power budget (10W) and lowest frequency setting for each module.

% Jay 11/3 : Commented following the merge below. See below subsubsection for more details.
% \subsubsection{Energy Consumption of Different Video Object Detection Models}
%
\noindent \textbf{Results}: Efficient object detector backbones (SSD, SSD+, EfficientDet D0, D0+, D3, D3+) consume lower instantaneous power measured in real-time. As shown in Figure~\ref{fig:energy}, the maximum peak power and the overall power level during inference is much lower up to 33\%, compared to other baselines.
%
% Main observation : lower power level & latency & energy consumption of + models.
Among all baselines, EfficientDet D0+, D3+, and SSD+ have a superior performance energy-wise, especially with SSD+ and EfficientDet D0+ both having less than 350 $J$. 
% Jay 11/3 : Added some more content.
Compared to its original detector backbone, EfficientDet D0+ on average consumes 290 $J$, where EfficientDet D0 consumes 1086 $J$, which is almost 4 times larger. The difference is bigger for EfficientDet D3+ and D3, where D3 consumes more than 5 times energy compared to D3+. SSD being a very light-weight detector backbone consumes only 722 $J$ on average, but still SSD+ is able to cut down the energy consumption down to 329 $J$ on average. 
%
% lower power level of efficient backbones compared to heavier baselines.
% Side observation
% Jay 11/3 : Reworded.
In addition, FRCNN+, and YOLO+ have low average total energy consumption at around 2,200 $J$, compared to FRCNN and YOLO. REPP with YOLOv3 consumed 3,346 $J$ on average while SELSA has the highest average energy consumption of 13,995 $J$, which is more than 10 times larger than adaptive baselines. Our overall energy consumption evaluations validate our insight that adaptive baselines EfficientDet D0+, D3+, SSD+, FRCNN+, and YOLO+ demonstrate superior energy efficiency relative to their rigid variants. 
%
% Side observation : oscillations
Since the major part of power consumption comes from the GPU module, and EfficientDet D0+, D3+ SSD+, FRCNN+, and YOLO+ which is our implementation leveraging an object tracker, we notice significant oscillations in their energy plots, compared to their original implementations (Fig.~\ref{fig:energy}). This is because the object tracker mainly uses the CPU and is more energy efficient. Switching between the object detector backbone (mainly executed on GPU) and object tracker (mainly executed on CPU) corresponds to the oscillations in the curve. 

We further investigate the impact of different power modes on the energy consumption of the 15 protocols.
We can see from Fig.~\ref{fig:energy} and Table~\ref{tab:energy} that all models in power mode 0 achieve lower latency than those in power modes 2 and 4, with the overall highest energy consumption. %This is because of its higher hardware frequency settings. 
All models in power mode 4 have the slowest inference latency with the lowest instantaneous power level compared with those in power modes 0 and 2. Particularly, compared with mode 0, power mode 4 helps reduce the instantaneous power on an average over models, by around 30\% and total energy consumption by around 10\% despite its longer average inference time of 18\%.

% Overall conclusion.
Overall, our efficiency knobs is able to cut down the energy consumption of object detection backbones significantly, showing at least 60\% decreased energy consumption for all adaptive baselines - EfficientDet D0+, D3+, SSD+, FRCNN+ and YOLO+ compared to its counterparts.
In addition, our evaluation on the energy consumption difference between different power modes suggests that the power mode can be utilized as another efficiency knob according to the user's requirements. Specifically, for video object detection tasks, the users can make their choice as to whether to focus on latency performance by choosing a higher performance power mode (\eg mode 0) or energy saving by switching to a lower performance power mode (\eg mode 4) of the embedded devices.

\section{Conclusion} \label{sec:conclusion}

% Jay 11/3 : Copied over, and created a new conclusion file.
% \somali{The conclusion has a few too many numerical results and it becomes rambling. Keep the full version as a comment and then extract out maybe 3/4 of the more important results to highlight.}

% Jay 11/3
\begin{comment}
Overall conclusion will be,
1. important method and implementation --> efficiency knobs / scheduler
2. evaluation
% 1. Overall performance of \name + different energy or latency requirements with \name.
% 2. Impact of power mode on energy vs. latency tradeoff
% 3. different devices, contention, multiple baselines for accuracy vs. latency
% 4. more in-depth analysis of energy.
try to keep the focus(or numerical results) for \name. Keep 3,4(benchmark section) relatively summarized.
\end{comment}

In this paper, we have proposed \name, an adaptive video object detection framework that consists of an object detector, object tracker, and a dynamic scheduler. A total of 8 different efficiency knobs are coupled with EfficientDet D0, D3, and SSD as the object detector backbones to create multiple execution branches. Further, our dynamic scheduler is able to predict the best performing branch at runtime.
% Evaluation #1 & 2
We evaluate \name from multiple perspectives, considering energy consumption, latency, and accuracy performance, alongside 15 different baselines.
Among all evaluated baselines, \name is able to achieve the best Pareto optimal performance curve, covering a wide spectrum of performance tradeoffs, with an inference time down to as low as 3.5 msec on the Xavier AGX board, and an accuracy up to 63.87\%. 
Moreover, \name is able to show dynamic switching of branches with flexibility to meet various user requirements. It is also observed that different power modes are able to provide further tradeoff of energy versus latency. For example, among our experimental settings of using AGX Xavier with power mode 0 and 2, we were able to achieve 40\% reduction in energy.

% Evaluation #3
We take one step further and perform a more in-depth evaluation of our multi-branch object detection kernels --- EfficientDet D0+, D3+ and SSD+ with the baselines on different runtime environment scenarios. Specifically, we evaluate using different embedded devices, different contention levels, and different power modes.
While all scenarios had impact on latency performance, our in-house adaptive baselines, coupled with our efficiency knobs, were able to meet certain user requirements at an acceptable accuracy. In contrast, all non-adaptive baselines suffer from a severe drop in latency performance, and more so, under GPU contention. 

% Evaluation #4
In the aspect of energy consumption, we show that using our efficiency knobs, our adaptive baselines --- EfficientDet D0+, D3+, SSD+, FRCNN+, and YOLO+ were at least 60\% more energy efficient compared to their non-adaptive counterparts. 
%
% SB (11/5/21): Repeated material and so chopped
% Our further evaluation with fine-grained power modes confirms that different power modes come with tradeoff between inference latency and energy consumption, and combined with our efficiency knobs, users may optimize further towards performance or efficiency matching their needs.

We hope that this work points to further work in understanding the suitability of various object detection kernels on embedded boards. This understanding must encompass varying levels of resource availability on these devices as well as varying power modes of operation available on these devices.

\bibliographystyle{ACM-Reference-Format}
\bibliography{7-reference}

\end{document}

\title{Supplementary of the Paper -- Benchmarking Video Object Detection Systems on Embedded Devices under Resource Contention}

\maketitle

\section{Something}

\begin{figure*}[t]
    \centering
    \begin{minipage}[c]{0.24\textwidth}
        \includegraphics[width=1\columnwidth]{figures/accuracy_p95_latency2_g0.png}
        \centerline{\small (a) No contention}
    \end{minipage}
    \hfill
    \begin{minipage}[c]{0.24\textwidth}
        \includegraphics[width=1\columnwidth]{figures/accuracy_p95_latency2_g20.png}
        \centerline{\small (b) 20\% GPU contention}
    \end{minipage}
    \hfill
    \begin{minipage}[c]{0.24\textwidth}
        \includegraphics[width=1\columnwidth]{figures/accuracy_p95_latency2_g50.png}
        \centerline{\small (c) 50\% GPU contention}
    \end{minipage}
    \hfill
    \begin{minipage}[c]{0.24\textwidth}
        \includegraphics[width=1\columnwidth]{figures/accuracy_p95_latency2_g90.png}
        \centerline{\small (d) 90\% GPU contention}
    \end{minipage}
    \caption{Benchmarking the object detection algorithms on the NVIDIA Jetson TX2.}
    \label{fig:accuracy_latency_TX2}
\end{figure*}

\begin{figure*}[t]
    \centering
    \begin{minipage}[c]{0.24\textwidth}
        \includegraphics[width=1\columnwidth]{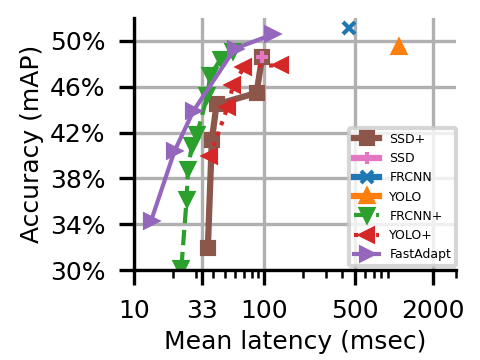}
        \centerline{\small (a) No contention}
    \end{minipage}
    \hfill
    \begin{minipage}[c]{0.24\textwidth}
        \includegraphics[width=1\columnwidth]{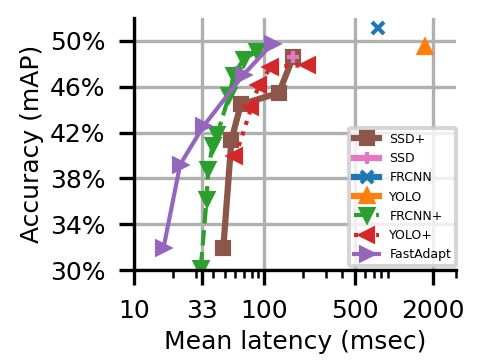}
        \centerline{\small (b) 20\% GPU contention}
    \end{minipage}
    \hfill
    \begin{minipage}[c]{0.24\textwidth}
        \includegraphics[width=1\columnwidth]{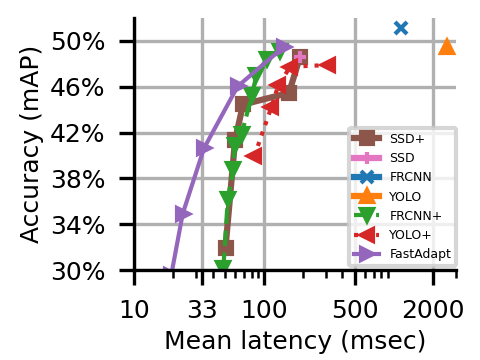}
        \centerline{\small (c) 50\% GPU contention}
    \end{minipage}
    \hfill
    \begin{minipage}[c]{0.24\textwidth}
        \includegraphics[width=1\columnwidth]{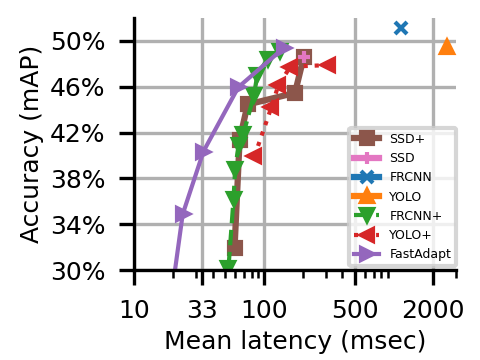}
        \centerline{\small (d) 90\% GPU contention}
    \end{minipage}
    \caption{Benchmarking the object detection algorithms on the NVIDIA Jetson Xavier NX.}
    \label{fig:accuracy_latency_NX}
\end{figure*}

\begin{figure*}[t]
    \centering
    \begin{minipage}[c]{0.24\textwidth}
        \includegraphics[width=1\columnwidth]{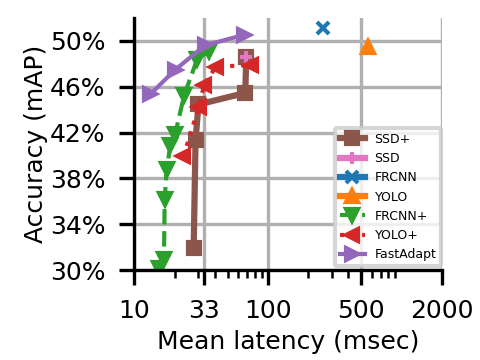}
        \centerline{\small (a) No contention}
    \end{minipage}
    \hfill
    \begin{minipage}[c]{0.24\textwidth}
        \includegraphics[width=1\columnwidth]{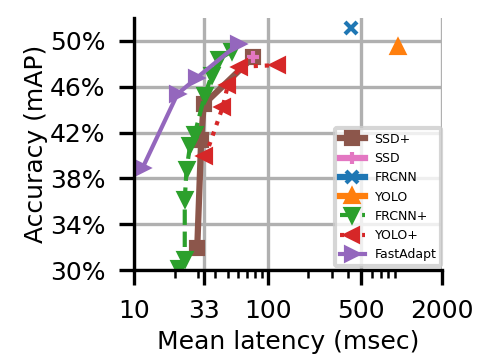}
        \centerline{\small (b) 20\% GPU contention}
    \end{minipage}
    \hfill
    \begin{minipage}[c]{0.24\textwidth}
        \includegraphics[width=1\columnwidth]{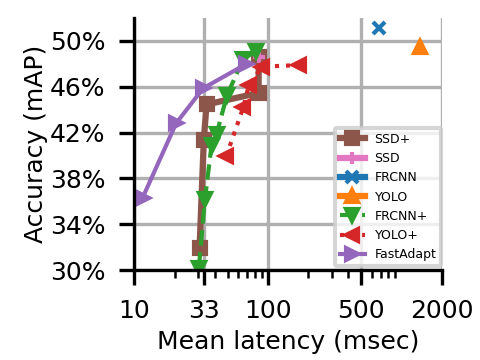}
        \centerline{\small (c) 50\% GPU contention}
    \end{minipage}
    \hfill
    \begin{minipage}[c]{0.24\textwidth}
        \includegraphics[width=1\columnwidth]{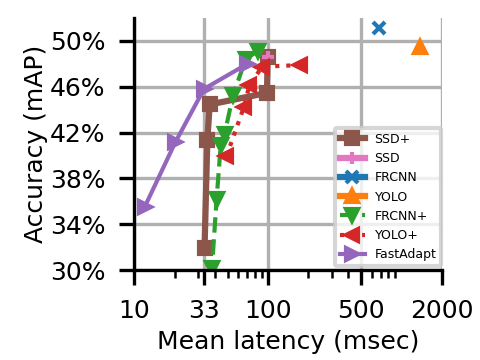}
        \centerline{\small (d) 90\% GPU contention}
    \end{minipage}
    \caption{Benchmarking the object detection algorithms on the NVIDIA Jetson AGX Xavier.}
    \label{fig:accuracy_latency_AGX}
\end{figure*}

\bibliographystyle{acm}
{\small \bibliography{6-reference}}